\pdfoutput=1

\documentclass[11pt]{article}

\usepackage[final]{acl}

\usepackage{times}
\usepackage{latexsym}

\usepackage[T1]{fontenc}

\usepackage[utf8]{inputenc}
\usepackage{amsmath}

\usepackage{microtype}
\usepackage{float}
\usepackage{inconsolata}
\usepackage{booktabs}
\usepackage{tabularx} 

\usepackage{graphicx}
\usepackage{makecell}
\usepackage{caption}
\usepackage{multirow}
\usepackage{enumitem}
\usepackage{booktabs} 
\title{Do Language Models Mirror Human Confidence? Exploring Psychological Insights to Address Overconfidence in LLMs}

\author{%
  \textbf{Chenjun Xu}$^1$\thanks{Equal contribution} \ \ \  \textbf{Bingbing Wen}$^1$\footnotemark[1] \\ 
  \ \ \  \textbf{Bin Han}\textsuperscript{1}\ \ \  \textbf{Robert Wolfe}\textsuperscript{1}  \ \ \ \textbf{Lucy Lu Wang}\textsuperscript{1,2} \ \ \ \textbf{Bill Howe}\textsuperscript{1} \\
\textsuperscript{1}University of Washington \ \ \ \ \ \textsuperscript{2}Allen Institute for AI \\
\texttt{\{bingbw,chenjux,bh193,rwolfe3,lucylw,billhowe\}@uw.edu} \ \ \ 
}

\begin{document}
\maketitle

\begin{abstract}

Psychology research has shown that humans are poor at estimating their performance on tasks, tending towards underconfidence on easy tasks and overconfidence on difficult tasks. We examine three LLMs, Llama-3-70B-instruct, Claude-3-Sonnet, and GPT-4o, on a range of QA tasks of varying difficulty, and show that models exhibit subtle differences from human patterns of overconfidence: less sensitive to task difficulty, and when prompted to answer based on different personas---e.g., expert vs layman, or different race, gender, and ages---the models will respond with stereotypically biased confidence estimations even though their underlying answer accuracy remains the same.  Based on these observations, we propose Answer-Free Confidence Estimation (AFCE) to improve confidence calibration and LLM interpretability in these settings. AFCE is a self-assessment method that employs two stages of prompting, first eliciting only confidence scores on questions, then asking separately for the answer. Experiments on the MMLU and GPQA datasets spanning subjects and difficulty show that this separation of tasks significantly reduces overconfidence and delivers more human-like sensitivity to task difficulty.\footnote{Code: https://github.com/chenjux/AFCE}

\end{abstract}

\section{Introduction}

\begin{figure}[t]
    \centering
 \includegraphics[width=0.45\textwidth]{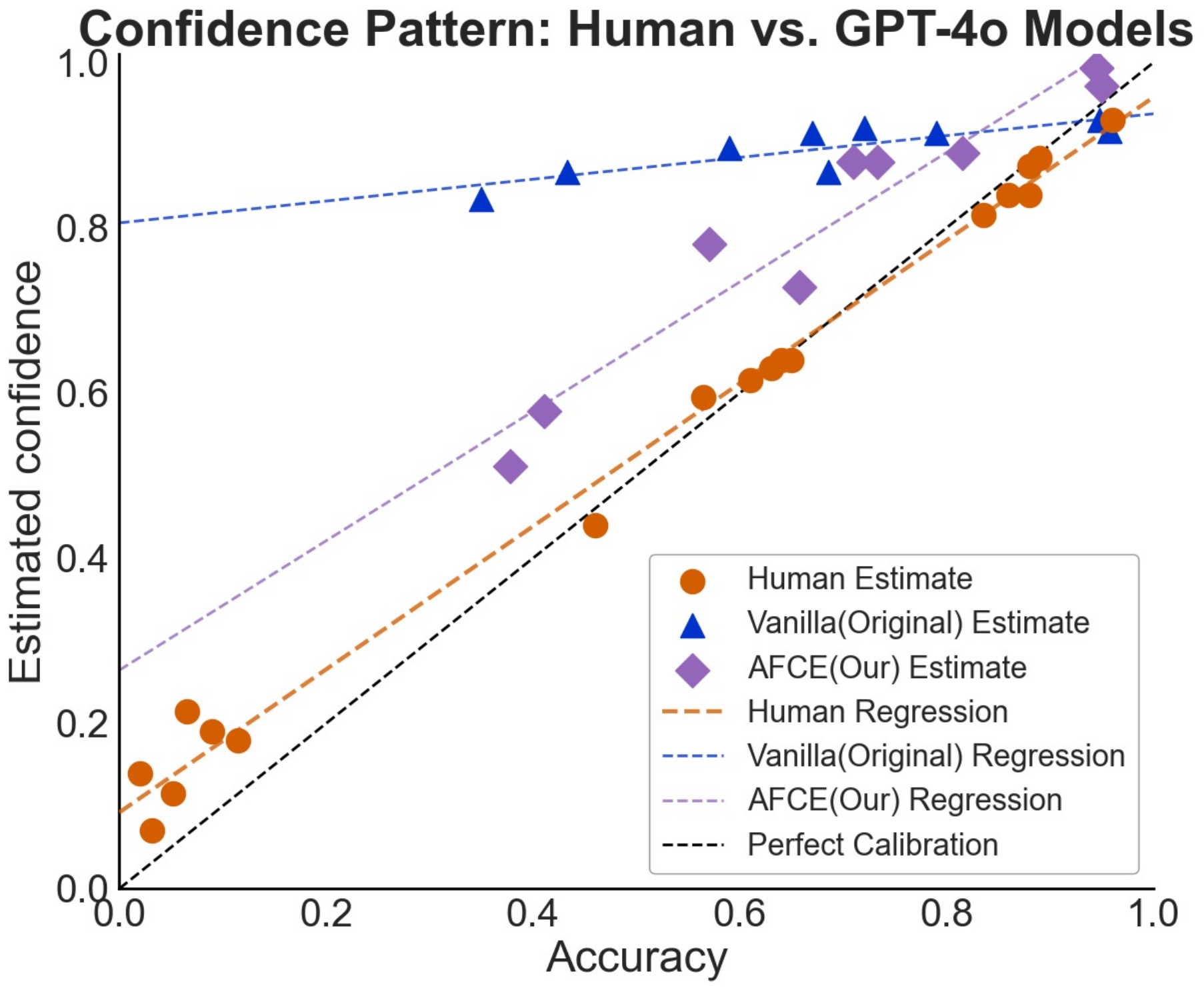} 
    \caption{ Comparative Analysis of Confidence Calibration Patterns Between GPT-4o and Human Participants. AFCE yields a steeper regression slope, demonstrating closer alignment with the optimal calibration line than existing methods. However, a notable disparity remains—GPT‑4o still shows less sensitivity to variation in task difficulty. Human data are from \citet{moore2008trouble}'s paper.}
    \label{fig:humanvsmodel} 
\end{figure}

Reliable confidence estimates are essential for effective human-machine collaboration~\cite{guo2017calibration}. Large language models (LLMs), however, are prone to overconfidence~\cite{xiong2024can}, which can result in inaccurate predictions when they should abstain~\cite{wen2024art}. As these models are increasingly deployed for real-world tasks such as medical diagnosis \cite{RosHoyo2024EvaluationOL}, legal analysis \cite{Deroy2024ApplicabilityOL}, and decision support systems \cite{enterprise_decision_support}, their performance directly impacts outcomes that affect human lives. Overconfidence in LLMs can lead to significant errors \cite{Zhou2023NavigatingTG}, reduced trust \cite{Kim2024ImNS}, and potentially harmful downstream consequences \cite{Li2023TheDS, wen-etal-2024-characterizing}. Prior work~\citep{10.5555/3618408.3618425, 10.1145/3586183.3606763} suggests that AI can reflect collective human-like behaviors, while also introducing new risks, such as amplification of misinformation. Therefore, understanding whether LLMs exhibit overconfidence in ways that parallel or exceed human patterns of overconfidence can inform improvements in reliability and safety in real-world applications. 


Human overconfidence is recognized as a significant cognitive bias~\cite{kruger1999unskilled}. \citet{moore2008trouble} reconcile experimental findings that individuals tend to (i) overestimate their own abilities on difficult tasks and underestimate them on easy tasks shown in Figure~\ref{fig:humanvsmodel}, and (ii) estimate that they outperform relative to others on easy tasks (overplacement) and underperform relative to others on hard tasks (underplacement). The authors~\citet{moore2008trouble} explain these phenomena using an information theoretic model demonstrating individuals' regressive estimates of their performance and even more regressive estimates of others' performance: on easy tasks, they underestimate their own success and others’ even more so; on hard tasks, they overestimate their own performance and others’ to an even greater degree.

Current confidence estimation approaches such as vanilla verbalized confidence reveals persistent overconfidence and highlights a disconnect between model-reported confidence and actual task performance as shown in Figure~\ref{fig:humanvsmodel}. Motivated by these observations, we propose Answer-Free Confidence Estimation (AFCE), which separates confidence estimation and answer generation into separate stages, and find that overconfidence effects as measured by Expected Calibration Error (ECE) are reduced in challenging tasks where overconfidence is pronounced. These results suggest that verbalized confidence methods should not assume human-like behavior in their design. 

We further use AFCE to explore whether confidence patterns observed in humans, as described by \citet{moore2008trouble}, are also present in LLMs. We uncover three distinct phenomena. (i) Models'  confidence scores are comparatively insensitive to task difficulty and exhibit only a weak correlation with actual accuracy, unlike human patterns reported by \citet{moore2008trouble}. (ii) When prompted with occupational personas, the model reflects stereotypical confidence levels (e.g., “layman” lower than “expert”) regardless of performance. (iii) Adding demographic cues (e.g., gender, race, age) further reduces expressed confidence, even when accuracy stays the same.

Our work investigates three core questions:
(i) Sensitivity to Task Difficulty: Is a model’s expressed confidence calibrated to task difficulty, and does it reflect the same over- and underestimation patterns observed in human judgment?
(ii) Over- and Underplacement Across Expertise: Do models demonstrate placement biases when estimating the performance of others, particularly when adopting personas with varying levels of expertise?
(iii) Demographic Bias in Confidence Expression: Do models exhibit systematic confidence biases when conditioned on demographic attributes such as race, gender, or age?

\noindent We summarize our contributions below:
\begin{itemize}[itemsep=-1pt, topsep=3pt, leftmargin=10pt]
    \item We evaluate LLMs' confidence estimations across tasks of varying difficulty. We observe that, similar to results in human subjects, LLMs exhibit underconfidence in task performance on easy tasks and overconfidence on hard tasks. However, model confidence estimates are less sensitive to task difficulty than human confidence estimates, suggesting a different mechanism mediates self-elicited confidence in LLMs. 
    
    \item We propose Answer-Free Confidence Estimation (AFCE), a method decoupling confidence estimation from answer generation to improve confidence calibration on challenging tasks and enables comparisons between human and LLM confidence patterns. We demonstrate the effectiveness of this method in three LLMs (LLaMa-3-70B, Claude-3-Sonnet, and GPT-4o). 

    \item We prompt the LLMs with personas from various levels of expertise to investigate over- and underplacement in LLMs and find that models consistently express lower confidence for ``Randomly chosen person'' persona or ``layman'' persona and higher confidence for ``expert'' persona, despite similar task accuracy. These results suggest verbalized confidence is influenced more by persona-based bias than actual performance.

    \item We assess how demographic personas influence the model’s confidence estimation. We find that LLMs tend to be underconfident when adopting \emph{any} human persona (with the exception of GPT-4o), but that the degree of underconfidence exhibits stereotypical biases: an Asian persona is more confident than other races; a female persona is less confident than other gender identities; a middle-aged persona tends to be more confident than other age groups.  These biases highlight the importance of considering demographic factors in confidence calibration, especially as role-playing techniques become more pervasive. 
\end{itemize}

\section{Related Work}\label{sec:related_work}
We review related work on human overconfidence and confidence elicitation methods for LLMs.

\textbf{Human Overconfidence} 
Overconfidence refers to an unjustified belief in one's knowledge and abilities \citep{kruger1999unskilled}, leading to undesirable outcomes in domains such as medicine \cite{Berner2008OverconfidenceAA}, politics \cite{Prooijen2021OverconfidenceIR}, and science~\citep{light2022knowledge}. 
Models to explain overconfidence have been broadly considered (e.g., Dunning-Kruger~\cite{kruger1999unskilled}, or recent results from \citet{sanchez2024intermediate} showing that those with intermediate knowledge may be the most overconfident).  In this paper, we focus on the experiments of \citet{moore2008trouble}, whose influential unifying model explained a variety of previous findings. 

In this study, we adopt the overconfidence measure from \citet{moore2008trouble} including over(under)estimation and over(under)placement.

\textbf{Confidence Elicitation in Language Models} Previous methods for eliciting confidence have primarily relied on white-box approaches, which have estimated confidence using token likelihoods~\cite{wang2024my} and internal state-based methods~\cite{kadavath2022language,kuhn2023semantic}. While effective, these techniques require internal access to the model, making them less applicable to models served over closed APIs, like GPT-4~\cite{achiam2023gpt}. Verbalized confidence approaches~\citep{tian-etal-2023-just}, primarily the vanilla method, appropriate to such models (\textit{i.e.,} prompting the model to produce confidence estimates in its output) tend to produce uniformly high estimations of model confidence, usually between 80\% and 100\% ~\cite{mielke2022reducing, xiong2024can}. To improve these estimates, some studies introduce consistency-based methods~\cite{lin2023generating, xiong2024can} to mitigate overconfidence. Other studies~\citep{kumar-etal-2024-confidence, tian-etal-2023-just} investigated the correlation between verbalized uncertainty and token probability and showed GPT-4o has strongest confidence-probability alignment across variety of tasks. 

In this study, we adopt these widely used confidence elicitation methods as baselines, but demonstrate a surprising divergence from human behavior that suggests a problematic decoupling of confidence estimation from answer generation. Indeed, prompting models to estimate confidence without producing answers reduces overconfidence and outperforms baseline methods on hard tasks.

\textbf{Roly-playing with Language Model} LLMs are increasingly employed to simulate human personas. Recent studies \citep{argyle2023out, aher2023using, park2022social, wen2024from} provide empirical evidence that LLM-driven simulations can replicate social science experiments and online forums with consistency comparable to data obtained from human participants. Likewise, \citet{wang-etal-2024-incharacter, jiang2023personallm} show that advanced role-playing agents exhibit personalities closely aligned with human perceptions, underscoring the effectiveness of role-playing approaches.

Such methodologies underscore the emergence of LLM-based role-play as a versatile, powerful tool across multiple domains \citep{10.1371/journal.pdig.0000205}. In this paper, we consider the interaction between role-playing and confidence estimation to investigate whether LLMs exhibit overplacement patterns akin to those observed in humans. We also address associated limitations and ethical considerations, motivating continued research into best practices for employing LLMs in role-playing contexts.

\textbf{Bias in LLM-Driven Computational Social Science} LLMs are known to poorly represent certain groups that make up a significant portion of the population (e.g., age 65+)~\citep{santurkar2023whose}. People perceive ChatGPT as predominantly male when asked about its gender~\citep{wan-etal-2023-kelly}, particularly about its core capabilities such as text summarization~\citep{wong2023chatgpt}. Moreover, ChatGPT has been shown to generate gender-biased responses~\citep{wan-etal-2023-kelly, hada-etal-2023-fifty}. Previous work~\citep{feng-etal-2023-pretraining} revealed that pretrained LLMs do have political leanings in pretraining corpora, propagating social biases into hate speech predictions and misinformation detectors. \citet{schramowski2022large} showed recent LLMs contain human-like biases about right and wrong behaviors, reflecting existing ethical and moral norms of society. 
\citet{dong2024not} confirmed a
substantial degree of in-group and out-group bias of LLMs across languages and personas. 

These latent (and in some applications, explicit) demographic roles can confound confidence estimation: roles associated with over- or underperformance in the training data are associated with verbalized over- or underconfidence estimates, even when actual performance is consistent.

\begin{figure}[t]
    \centering
    \includegraphics[width=0.5\textwidth]{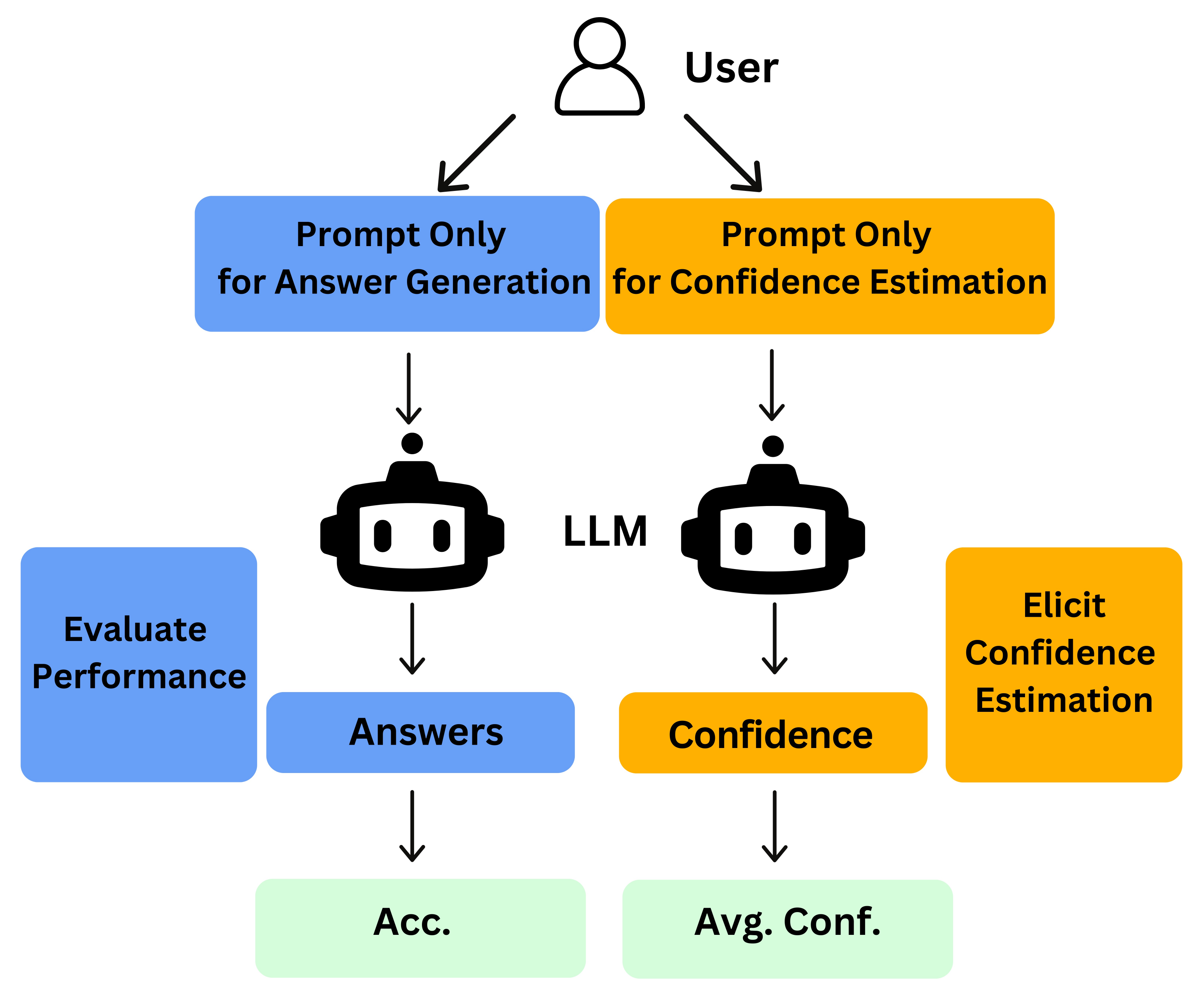}
    \caption{ AFCE gathers confidence levels for a set of questions without requiring answers, thereby separating confidence estimation from the answering process. We utilize this approach to simulate human test confidence in psychology and compare the confidence patterns of humans and LLMs.}
    \vspace{-3mm}
    \label{fig:AFCE}
\end{figure}


\section{Data \& Models}


\textbf{Datasets} Following the 
experiment design of \citet{moore2008trouble},\footnote{We are unable to repeat the exact questions used in ~\citet{moore2008trouble}, as they were not available.} which examined question banks spanning various difficulty levels (high school, college, expert) and subjects (physics, chemistry, biology), we use two datasets: 
\begin{itemize}[noitemsep, topsep=0pt, leftmargin=10pt]
    \item MMLU \cite{hendrycks2021measuring}, a collection of domain-specific multiple-choice questions across 57 subjects at multiple educational stages. We select MMLU questions from  High School and College in the subject areas of Physics, Chemistry, and Biology. 
    \item GPQA \cite{rein2023gpqa}, a dataset of multiple-choice questions created by experts (\textit{i.e.}, individuals holding or pursuing a PhD). We select questions from physics, chemistry, and biology to represent expert-level difficulty. 
\end{itemize}
Each (subject, difficulty) pair is considered a distinct subtask. To afford a meaningful comparison, we calculate accuracy and confidence in a manner consistent with \citet{moore2008trouble}, where 10 questions are evaluated in each prompt. Further dataset details can be found in Appendix Table~\ref{tab:dataset_examples}.

\begin{figure*}[t]
    \centering
    \includegraphics[width=1\textwidth]{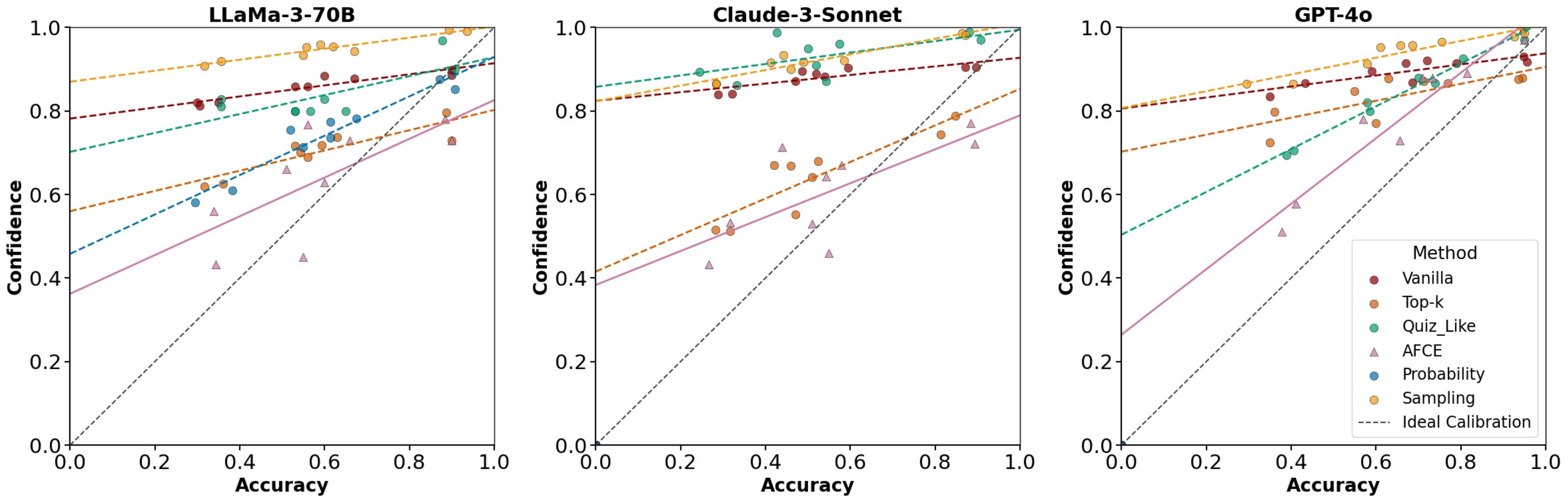} 
    \caption{AFCE reduces overconfidence across models and improves sensitivity to task difficulty for GPT-4o. For GPT-4o, AFCE produces a steeper regression slope that aligns more closely with the ideal calibration line. For Claude-3, AFCE and top-k perform comparably, both exceeding other methods. The relatively lower regression line for AFCE for all models suggests reduced overconfidence.}
    \label{fig:compare_method} 
\end{figure*}

\textbf{Models} 
We examine the widely used open-source LLM Meta-Llama-3-70B-Instruct (LLaMA-3-70B), as well as two API-based models, Claude-3-sonnet-20240229 (Claude-3) and GPT-4o-2024-05-13 (GPT-4o). To enhance reproducibility and limit variability in model outputs, we set temperature to 0 and top-$p$ sampling to 1. Model details are provided in Appendix Table~\ref{tab:model_statistics}. Additional experiments on Gemma and Mistral, along with their results, are presented in Appendix Figures~\ref{fig:overplacement_combined} and \ref{fig:bias_results_1}.

\begin{table*}[t!]
    \centering
    \scriptsize
    \setlength{\tabcolsep}{2.5pt}
    \begin{tabular}{@{}l*{18}{|c}|c@{}}
        \toprule
        \multirow{3}{*}{\textbf{Method}} & \multicolumn{6}{c|}{\textbf{High School}} & \multicolumn{6}{c|}{\textbf{College}} & \multicolumn{6}{c|}{\textbf{Expert}} & \multirow{3}{*}{\textbf{AvE}} \\
        \cmidrule(lr){2-7} \cmidrule(lr){8-13} \cmidrule(lr){14-19}
        & \multicolumn{2}{c|}{Physics} & \multicolumn{2}{c|}{Chemistry} & \multicolumn{2}{c|}{Biology} & \multicolumn{2}{c|}{Physics} & \multicolumn{2}{c|}{Chemistry} & \multicolumn{2}{c|}{Biology} & \multicolumn{2}{c|}{Physics} & \multicolumn{2}{c|}{Chemistry} & \multicolumn{2}{c|}{Biology} & \\
        \cmidrule(lr){2-7} \cmidrule(lr){8-13} \cmidrule(lr){14-19}
        & Acc & ECE & Acc & ECE & Acc & ECE & Acc & ECE & Acc & ECE & Acc & ECE & Acc & ECE & Acc & ECE & Acc & ECE & \\
        \midrule
        \multicolumn{20}{c}{\textbf{LLaMA-3-70B}} \\
        \midrule
        Vanilla & 60.0 & 28.4 & \underline{67.0} & 20.7 & \textbf{90.0} & \textbf{2.1} & 53.0 & 32.8 & \textbf{56.0} & 29.9 & 90.0 & \textbf{1.4} & 35.0 & 47.0 & 30.6 & 50.7 & 54.3 & 29.9 & 27.0 \\
        Top-K & 59.3 & \textbf{12.7} & 63.0 & \underline{12.2} & 88.7 & 9.3 & \underline{56.0} & 16.8 & 53.0 & \underline{19.1} & 87.9 & 10.8 & \underline{36.1} & 26.4 & 31.7 & 32.0 & 54.3 & 15.7 & 17.9 \\
        Sampling & \textbf{62.0} & 33.4 & \underline{67.0} & 27.8 & \underline{89.4} & 10.0 & \textbf{59.0} & 36.9 & 55.0 & 41.0 & \textbf{93.6} & 5.5 & 35.6 & 56.4 & 31.7 & 59.2 & 55.7 & 39.5 & 34.4 \\
        Probability & \underline{61.3} & 17.7 & \textbf{67.5} & 11.5 & 87.1 & 5.2 & 55.0 & \underline{16.4} & 52.0 & 23.6 & 90.7 & 7.7 & \textbf{38.3} & \underline{25.5} & 29.4 & \underline{28.6} & \textbf{61.4} & \underline{14.2} & \underline{15.1} \\
        Quiz & 56.7 & 23.3 & 65.0 & 15.0 & 87.7 & \underline{9.0} & 53.0 & 27.0 & 53.0 & 27.0 & \underline{90.7} & \underline{5.0} & 35.6 & 45.6 & \textbf{35.6} & 47.2 & \underline{60.0} & 22.9 & 24.7 \\
        AFCE & 56.0 & \underline{20.7} & 66.0 & \textbf{11.0} & 88.4 & 10.3 & 51.0 & \textbf{15.0} & \underline{55.0} & \textbf{6.0} & 90.0 & 17.1 & 34.4 & \textbf{16.7} & \underline{33.9} & \textbf{22.2} & \underline{60.0} & \textbf{11.4} & \textbf{14.5} \\
        \midrule
        \multicolumn{20}{c}{\textbf{Claude-3-sonnet}} \\
        \midrule
        Vanilla & \textbf{48.7} & 40.9 & \textbf{59.5} & 30.9 & \textbf{89.7} & \textbf{2.3} & \textbf{52.0} & 37.0 & \underline{54.0} & 35.9 & 87.1 & \textbf{4.2} & \underline{28.9} & 55.6 & \underline{32.2} & 51.8 & 47.1 & 39.9 & 23.2 \\
        Top-K & 42.0 & \textbf{25.2} & 52.5 & \underline{15.5} & 84.8 & \underline{6.9} & 46.0 & \underline{20.9} & 51.0 & \underline{13.2} & 81.4 & 8.1 & \textbf{31.7} & \underline{19.5} & 28.3 & \underline{24.4} & 47.1 & \underline{12.6} & 21.7 \\
        Sampling & 41.3 & 50.2 & 58.5 & 33.7 & 87.1 & 11.5 & 46.0 & 44.0 & 49.0 & 42.7 & 86.4 & 12.2 & 28.3 & 62.3 & 28.3 & 58.2 & 44.3 & 49.1 & 29.0 \\
        Quiz & 42.7 & 56.0 & 57.5 & 38.5 & 88.1 & 11.0 & 50.0 & 45.0 & 52.0 & 39.0 & \textbf{90.7} & \underline{7.9} & 24.4 & 65.0 & \textbf{33.3} & 52.8 & 54.3 & 32.9 & \underline{17.9} \\
        AFCE & \underline{44.0} & \underline{27.3} & \underline{58.0} & \textbf{9.0} & \underline{88.4} & 11.3 & \underline{51.0} & \textbf{2.0} & \textbf{55.0} & \textbf{9.0} & \underline{89.3} & 17.1 & 26.7 & \textbf{16.7} & 31.7 & \textbf{21.7} & \textbf{54.3} & \textbf{10.0} & \textbf{13.2} \\
        \midrule
        \multicolumn{20}{c}{\textbf{GPT-4o}} \\
        \midrule
        Vanilla & 72.0 & 20.6 & 79.0 & 12.5 & \textbf{94.8} & \textbf{2.9} & 67.0 & 25.3 & \textbf{59.0} & 31.1 & \textbf{95.7} & \underline{4.0} & \textbf{43.3} & 43.4 & 35.0 & 48.5 & \textbf{68.6} & 20.1 & 33.2 \\
        Top-K & 71.3 & 15.7 & 77.0 & \underline{10.9} & 94.5 & 6.7 & 63.0 & 26.7 & 55.0 & 30.6 & 93.6 & 6.0 & 36.1 & 44.3 & 35.0 & 37.4 & 60.0 & \underline{17.1} & \underline{16.3} \\
        Sampling & 68.7 & 28.0 & 75.5 & 21.7 & 92.6 & 5.3 & 61.0 & 34.3 & \underline{58.0} & 35.2 & 95.0 & \textbf{3.6} & 40.6 & 45.9 & 29.4 & 57.0 & 65.7 & 30.0 & 40.4 \\
        Quiz & \textbf{74.0} & \textbf{12.7} & \underline{80.5} & 12.0 & \textbf{94.8} & \underline{4.8} & \underline{70.0} & \underline{18.0} & \underline{58.0} & \underline{24.0} & \underline{95.0} & 5.0 & 40.6 & \underline{32.2} & \textbf{38.9} & \underline{30.6} & 58.6 & 21.4 & 38.7 \\
        AFCE & \underline{73.3} & \underline{14.7} & \textbf{81.5} & \textbf{9.5} & 94.5 & 6.1 & \textbf{71.0} & \textbf{17.0} & 57.0 & \textbf{21.0} & \underline{95.0} & \textbf{3.6} & \underline{41.1} & \textbf{21.1} & \underline{37.8} & \textbf{16.1} & \underline{65.7} & \textbf{10.0} & \textbf{13.8} \\
        \bottomrule
    \end{tabular}
    \caption{Confidence elicitation and performance comparison across models, subjects and difficulty levels. AFCE significantly reduces overconfidence and achieves better calibration performance compared to other baseline methods, especially for challenging tasks. Acc: Accuracy, ECE: Expected Calibration Error. AvE: Average ECE. All values are percentages.}
    \label{tab:method_compare_all}
\end{table*}

\section{Over(Under) Estimation Across Task Difficulty}


In this section, we study the relationship between the LLMs' confidence estimation and task difficulty. We compare our Answer-Free Confidence Estimation (AFCE) method with other commonly used confidence estimation methods. 


\subsection{Experiment Setup}

\textbf{Baselines}. We consider five baseline methods for model confidence estimation: three verbalized approaches that prompt the model to express confidence directly; one sampling-based method that infers confidence from output variability across generations; and one probability-based method that uses the token probability as confidence proxy.
\begin{itemize}[noitemsep, topsep=0pt, leftmargin=10pt]
\item Vanilla Verbalized Confidence~\cite{lin2022teaching}, which prompts the model with "Read the question, provide your answer, and report your confidence in this answer".
\item Top-$k$ Prompting Verbalized Confidence~\cite{tian-etal-2023-just}, which prompts the model to provide "your K best guesses and the probability that each option is correct (0\% to 100\%) for the following question".
\item Quiz-Like Prompting inspired by ~\citet{moore2008trouble}, which prompts the model to "Answer the following 10 questions and estimate how many were answered correctly". 
\item Sampling-based. Sampling strategy refers to self-random sampling strategy (with three samples) combined with Avg-Conf Aggregation, as proposed by ~\cite{xiong2024can}.
\item Probability-based. We use the probability assigned to the first token as confidence~\citep{wang2024my}.

\end{itemize}

\begin{figure*}[t]
    \centering
    \includegraphics[width=\textwidth]{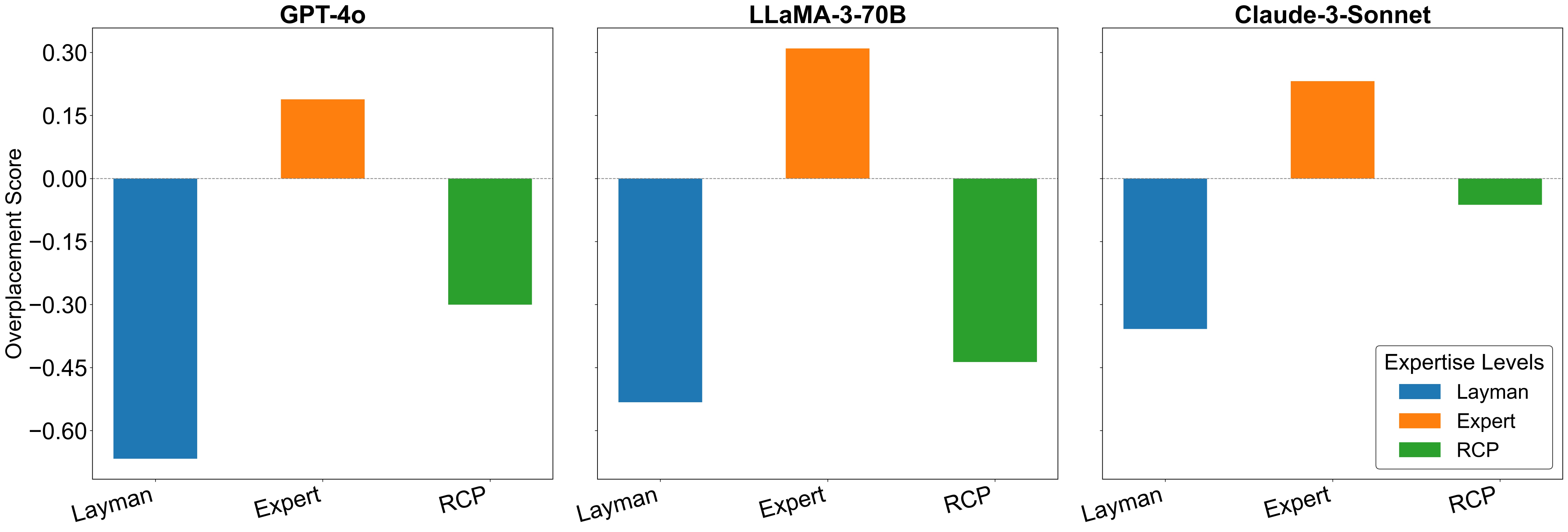}
    
    \caption{\textbf{Overplacement Score} quantifies the degree of overplacement, calculated as:
$(\text{Confidence}_{\text{Estimate Others}} - \text{Accuracy}_{\text{Others}}) - (\text{Confidence}_{\text{Self-Estimate}} - \text{Accuracy}_{\text{Self}})$.
All models exhibit overplacement towards  \textit{Expert} persona and underplacement towards \textit{Layman} persona. GPT-4o and LLaMA-3-70B also show underplacement in the \textit{Randomly Chosen Person} role, whereas Claude-3 demonstrates better calibration.}
    \label{fig:overplacement_result_score}
\end{figure*}

\textbf{Our Method.} We propose \textit{Answer-Free Confidence Estimation (AFCE)}, which distinguishes task performance evaluation from  confidence estimation shown in Figure~\ref{fig:AFCE}.  To evaluate performance, we prompt the model with "Please answer the following 10 questions by selecting only the option letter," and we use the model's responses to compute its accuracy. We separately obtain the model's confidence by prompting the model to "Read the questions and estimate how many you can answer correctly (choose a number from 0-10)." We hypothesize that task performance and confidence estimation are mediated by different mechanisms, such that task execution can confound confidence estimation, leading to overconfidence. We include the prompt template in Appendix Table~\ref{tab:hardness_prompts}.

\textbf{Evaluation} We use Expected Calibration Error (ECE)~\cite{Guo2017OnCO} with 10 bins to evaluate confidence calibration, which quantifies the difference between the confidence and actual accuracy. 

\vspace{-2mm}
\subsection{Results}


\textbf{AFCE can produce confidence estimation that is sensitive to task difficulty in GPT-4o.} Figure~\ref{fig:compare_method} shows that GPT-4o’s calibration slope becomes significantly steeper relative to the ideal line when applying our method, indicating enhanced sensitivity despite not achieving perfect calibration. We interpret difficulty through two lenses: (1) stated education level (high school, college, expert) and (2) actual task performance. While LLaMA-3-70B and Claude-3 exhibit relatively flat confidence curves regardless of accuracy, GPT-4o’s confidence estimation is more sensitive to performance. This flatness suggests that LLaMA-3-70B and Claude-3 rely on a “standard” confidence level, limiting the effectiveness of verbalized confidence-elicitation strategies. Figure~\ref{fig:hardness} in the Appendix further illustrates the relationship between confidence estimation and (subject, difficulty).


\textbf{AFCE consistently outperforms other baseline approaches in calibration performance across models on challenging tasks}.  As shown in Table~\ref{tab:method_compare_all}, we found while LLMs exhibit lower accuracy on expert-level tasks, they sometimes exhibit stronger performance on college-level subjects than on high school-level subjects, breaking with typical human judgments of task difficulty. As the difficulty increases---particularly at the expert level---AFCE demonstrates consistently stronger improvements in ECE compared to all baselines. Table~\ref{tab:method_compare_all} illustrates this trend: the progression from high school to college to expert difficulty highlights bold (i.e., best) results for AFCE at higher levels. For GPT-4o, applying AFCE decreases average ECE by 58.4\% compared to the vanilla prompt method, 63.8\% relative to the quiz-like prompt, 65.8\% against the sampling-based method, and 15.3\% compared to top-$k$ prompts, outperforming all other baselines.

 
Figure~\ref{fig:compare_method} demonstrates that the AFCE method substantially alleviates overconfidence across models. The corresponding regression line for the AFCE approach (purple line) is consistently lower than those associated with other methods. We posit that task-insensitive overconfidence in the vanilla case results from the epistemically intensive process of generating factual information~\citep{teplicasciurus} dominating the reasoning, leading the models to default to a typical confidence answer regardless of question difficulty. 
But when the tasks are separated, the model is able to fully attend to confidence estimation and become more accurate, more sensitive to task difficulty, and therefore more human-like.  
Additionally, omitting answer generation may prevent the model from ``overthinking''~\citep{cuadron2025danger} which may lead to overconfidence. We intend to investigate these underlying mechanisms and their relationship to human cognition in future work aimed at expanding the utility of AFCE.

Given that AFCE outperforms baselines, it will be employed in the subsequent overplacement and demographic bias experiments.

\section{Over(Under)placement Across Levels of Expertise} 
\label{overplacement}
In psychology, overplacement is similar to overconfidence but  refers to an inaccurate belief about one’s abilities, performance, or qualities \textit{compared to others}, often overestimating one’s relative standing~\citep{moore2008trouble}. 



\subsection{Experiment Setup} 


\begin{figure*}[t]
    \centering
    \includegraphics[width=0.85\textwidth]{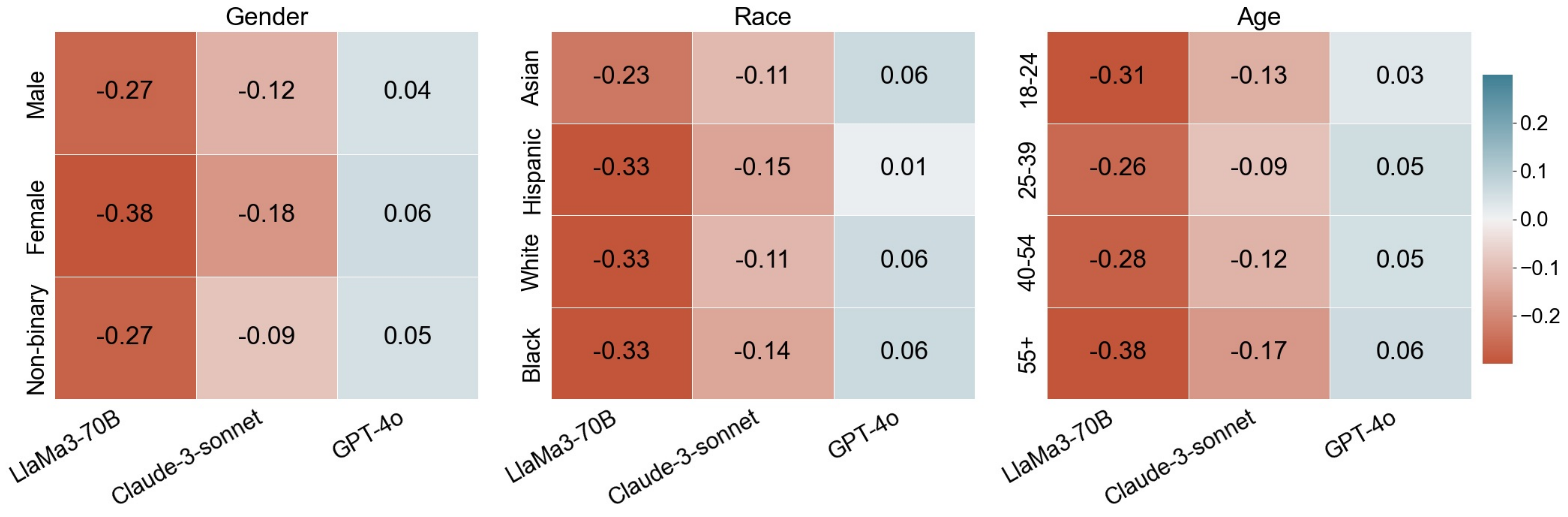} 
    \caption{$\Delta_\text{Demographic}$ measures the gap between confidence and accuracy (Confidence-Accuracy) across various demographic groups (gender, race, or age). GPT-4o demonstrates more balanced confidence estimations while LLaMA-3-70B and Claude-3 show consistent underconfidence across all demographic groups. }
    \label{fig:bias_results}
\end{figure*}

We adapt the experiments on human subjects ~\cite{moore2008trouble} for LLMs by prompting language models to adopt the personas of other individuals to estimate comparative confidence. Specifically, We prompt the model to adopt the persona with a particular expertise level then to 1) answer the questions and 2) estimate its confidence using AFCE. We specifically instruct the model to adopt the persona of a ``randomly chosen person'', an ``expert'' in the subject under consideration, and a ``layman'' with regard to the subject. In prompting language models to adopt personas~\citep{shanahan2023role, wang2023rolellm, hagendorff2023machine, shah2023scalable}, we follow recent work on using LLMs for simulation in computational social science \cite{aher2023using}, as well as assessments of model bias and fairness \cite{Cheng2023MarkedPU}. The template for overplacement prompts is provided in Appendix Table~\ref{tab:overplacement_prompts}. We measure overplacement as the difference between over(under)confidence in others and over(under)confidence in self (see caption of Figure \ref{fig:overplacement_result_score}). 



\subsection{Results} 

\textbf{LLMs exhibit substantial overplacement toward expert personas, while demonstrating notable underplacement toward average individuals and layperson personas.} Figure~\ref{fig:overplacement_result_score} compares the extent of overplacement across different expertise personas for LLaMA-3-70B, GPT-4o, and Claude-3. All models consistently show pronounced overplacement for expert roles, contrasted with clear underplacement for average and layperson roles. Additionally, models differ significantly in how they assess the performance of a randomly selected individual: LLaMA-3-70B and GPT-4o display marked underplacement biases against this neutral baseline, whereas Claude-3 provides relatively balanced estimates. Despite these overplacement biases, actual model accuracy remains relatively consistent across different personas, indicating a systematic tendency to overestimate confidence in expert roles and underestimate confidence in layperson roles. Overall, the models' confidence judgments about others seem disconnected from their true capabilities when adopting persona-based prompts. We expect that this mismatch could prove problematic for the range of research that now utilizes persona-prompted LLMs in social scientific simulations \cite{aher2023using,Ziems2023CanLL}.

\section{Bias in Confidence Estimation Across Demographic Personas}\label{sec:bias}
Here, we combine the AFCE method and role-playing 
to analyze variations in confidence estimation across different protected attributes (race, gender, and age).


\subsection{Experiment Setup} 

We utilize LLMs within a structured role-playing framework (following the setup described in \S\ref{overplacement}) to simulate individuals from various demographic groups as follows:

\small
\[
\begin{cases}
\text{Race} \in \{\text{White}, \text{Black}, \text{Asian}, \text{Hispanic}\} \\
\text{Gender} \in \{\text{Male}, \text{Female}, \text{Non-binary}\} \\
\text{Age} \in \left\{
  \begin{aligned}
  &\text{Young adult (18--24)}, \; \text{Adult (25--39)},\\
  &\text{Middle-aged (40--54)}, \; \text{Senior (55+)}
  \end{aligned}
\right\}
\end{cases}
\]
\normalsize
Overplacement prompts are provided in Appendix Table~\ref{tab:demographic_prompts}.

\subsection{Results}

\textbf{LLMs tend to be underconfident when adopting \emph{any} human persona (with the exception of GPT-4o), but that the degree of underconfidence exhibits stereotypical biases.} Figure~\ref{fig:bias_results} illustrates the accuracy and confidence estimates of LLMs when role-playing different assignments of gender, race, and age. Variations in $\Delta_\text{Demographic}$ are mainly driven by confidence estimation, as prediction differences remain minimal, suggesting independent confidence estimation and question-answering mechanisms.

LLaMA3-70B and Claude-3 underestimate confidence when playing female-identifying roles, with $\Delta_\text{Demographic}$ ranking as Non-binary $\geq$ Male $>$ Female. GPT-4o demonstrates nearly uniform $\Delta_\text{Demographic}$ values, indicating reduced gender biases. Under racial assignments, LLaMA3-70B shows larger negative $\Delta_\text{Demographic}$ for Asians, while Claude-3 exhibits higher negative values for White and Asian individuals. GPT-4o displays smaller disparities, indicating more balanced racial confidence estimations. For age groups, LLaMA3-70B and Claude-3 show underestimation for 18–24 and 55+ groups but favor 25–39 and 40–54 cohorts while GPT-4o maintains minimal bias.

The LLaMA and Claude models show consistent underconfidence across all demographic groups, with lower negative $\Delta_\text{Demographic}$ values, reflecting a misalignment between their confidence levels and actual performance. Despite lower confidence, their accuracy often remains unaffected, especially in complex demographic contexts, highlighting calibration issues. In contrast, GPT-4o demonstrates balanced confidence estimations across races, ages, and genders, effectively mitigating biases. Its consistent performance makes it well-suited for research requiring fairness and inclusivity, paving the way for equitable insights in social science studies.

\begin{table}[t]
    \centering
    \small
    \begin{tabular}{l r r r}
        \toprule
        \textbf{NQ-open} & \textbf{Acc} & \textbf{AvC} & \textbf{ECE} \\
        \midrule
        Quiz-Like & 74.0 & 78.0 & 6.0 \\
        Vanilla  & 74.0 & 77.2 & 6.0 \\
        AFCE     & 74.0 & 75.0 & \textbf{4.0} \\
        \midrule
        \textbf{SimpleQA} & \textbf{Acc} & \textbf{AvC} & \textbf{ECE} \\
        \midrule
        Quiz-Like & 36.0 & 78.0 & 42.0 \\
        Vanilla  & 31.0 & 87.0 & 56.0 \\
        AFCE     & 36.0 & 25.0 & \textbf{6.0} \\
                \bottomrule
    \end{tabular}
    \caption{AFCE’s performance on open-ended generation QA using GPT-4o indicates that its calibration surpasses that of other verbalized baselines. Acc: Accuracy, AvC: Average Confidence, ECE: Expected Calibration Error. All values are percentages.}
    \label{tab:open_question}
\end{table}

\section{Ablation Study}

We investigate whether AFCE generalizes to open-ended QA and whether it remains robust to variations in question order and the number of questions within a prompt. Accordingly, we conducted the following two experiments using GPT-4o.

\textbf{AFCE Generalizes to Open-Ended Generation QA.} We randomly sampled 100 open-ended questions each from NaturalQuestions-open \cite{10.1145/3586183.3606763} and SimpleQA \cite{wei2024measuringshortformfactualitylarge}. Experimental results presented in Table~\ref{tab:open_question} indicate that AFCE consistently outperforms other baselines across both datasets. Although our primary focus was on multiple choice questions for cross-domain comparisons and alignment with human subject studies, these findings suggest that AFCE extends beyond structured formats.


\textbf{AFCE Demonstrates Robustness Under Variations in Question Order and Group Size.} We conducted two experiments: (1) reducing the group size to five questions, and (2) randomizing question order. We observed no substantial differences in performance under either condition. As shown in Table~\ref{tab:random}, AFCE maintains its effectiveness across these configurations.

\section{Discussion}
Our study provides insights into LLM confidence calibration informed by the psychology literature, exposing subtle differences between LLM and human behavior with implications for the understanding of confidence estimation and the design of confidence calibration techniques.

 While our method AFCE effectively mitigates overconfidence in challenging tasks, it may lead to underconfidence in easier tasks. This effect could stem from the decoupling of answer generation from confidence estimation. However, the underlying mechanism by which this decoupling influences confidence estimation remains unclear. It is possible that AFCE adjusts confidence more aggressively for challenging questions, while its influence is attenuated for simpler ones. Further investigation is needed to clarify how and why this decoupling shapes confidence across varying task difficulties.

We find that LLMs can be easily manipulated by expert personas, leading to inflated confidence scores. This highlights key directions for future work: first, understanding how confidence signals influence human trust and decision-making in real-world settings; second, developing dynamic confidence calibration methods tailored to specific applications or user profiles to better manage risk.
Bias in confidence calibration across demographics, such as race, gender, and age, poses additional concerns. While recent models like GPT-4o show progress in reducing these biases, other LLMs still exhibit notable disparities. This warrants further investigation, as addressing confidence calibration alone may be insufficient to ensure fairness and reliability. Further experiments should explore how demographic-aware verbalized confidence estimation can help mitigate bias across different groups.

\section{Conclusion}

We analyze LLM overconfidence from a psychological perspective, showing that LLMs are less sensitive to task difficulty and can drastically shift their confidence when adopting personas—despite minimal changes in actual performance. This suggests that answer generation and confidence estimation involve distinct processes differently influenced by prompts and biases. We propose AFCE, which significantly reduces ECE for difficult reasoning questions prone to overconfidence. We also find that most LLMs (except GPT-4o) become underconfident when adopting Asian, age 55+, or female personas, again without affecting task performance. Likewise, a model’s self-assessment is typically overconfident, yet it claims higher confidence as an expert and lower confidence as a novice, with no notable impact on performance.

\section{Limitations}

 In this study, we mainly focus on verbalized confidence elicitation methods which are accessible to all kinds of models. Our investigation shows a promising correlation between a model's confidence pattern and humans' confidence pattern, but heavily depends on precise prompting techniques. Besides, although our method outperforms in the ECE metric, ECE has limitations, including its sensitivity to bin definitions and its inability to account for the overall prediction distribution. The analysis was conducted on a select group of LLMs: GPT-4o, Claude-3, LLaMA-3, Mistral and Gemma. These models were chosen for their architectural diversity and representativeness of current state-of-the-art. However, the inclusion of other models, such as Gemini Pro, PerplexityAI or those specialized in specific languages and domains, in future studies would likely reveal further interesting findings.  


\section{Ethical Considerations}
This study highlights key ethical concerns in LLM confidence calibration. Disparities in confidence estimation across demographics, such as race, gender, and age, risk perpetuating inequities, particularly in sensitive areas like healthcare, education, and law. Overconfidence, especially in "Expert" roles, may lead to unwarranted trust in incorrect outputs, requiring careful prompt design to mitigate risks. Transparency is essential to communicate model limitations and build user trust, while biases and confidence misalignment may spread misinformation or disproportionately affect marginalized groups. Addressing these issues is crucial for fairness, reliability, and ethical LLM deployment.

\section*{Acknowledgements}

This work is partially supported by gift funds from the Allen Institute for AI.

\newpage

\begin{thebibliography}{68}
\providecommand{\natexlab}[1]{#1}

\bibitem[{Achiam et~al.(2023)Achiam, Adler, Agarwal, Ahmad, Akkaya, Aleman, Almeida, Altenschmidt, Altman, Anadkat et~al.}]{achiam2023gpt}
Josh Achiam, Steven Adler, Sandhini Agarwal, Lama Ahmad, Ilge Akkaya, Florencia~Leoni Aleman, Diogo Almeida, Janko Altenschmidt, Sam Altman, Shyamal Anadkat, et~al. 2023.
\newblock Gpt-4 technical report.
\newblock \emph{arXiv preprint arXiv:2303.08774}.

\bibitem[{Aher et~al.(2023{\natexlab{a}})Aher, Arriaga, and Kalai}]{10.5555/3618408.3618425}
Gati Aher, Rosa~I. Arriaga, and Adam~Tauman Kalai. 2023{\natexlab{a}}.
\newblock Using large language models to simulate multiple humans and replicate human subject studies.
\newblock In \emph{Proceedings of the 40th International Conference on Machine Learning}, ICML'23. JMLR.org.

\bibitem[{Aher et~al.(2023{\natexlab{b}})Aher, Arriaga, and Kalai}]{aher2023using}
Gati~V Aher, Rosa~I Arriaga, and Adam~Tauman Kalai. 2023{\natexlab{b}}.
\newblock Using large language models to simulate multiple humans and replicate human subject studies.
\newblock In \emph{International Conference on Machine Learning}, pages 337--371. PMLR.

\bibitem[{Argyle et~al.(2023)Argyle, Busby, Fulda, Gubler, Rytting, and Wingate}]{argyle2023out}
Lisa~P Argyle, Ethan~C Busby, Nancy Fulda, Joshua~R Gubler, Christopher Rytting, and David Wingate. 2023.
\newblock Out of one, many: Using language models to simulate human samples.
\newblock \emph{Political Analysis}, 31(3):337--351.

\bibitem[{Bengio and LeCun(2007)}]{Bengio+chapter2007}
Yoshua Bengio and Yann LeCun. 2007.
\newblock Scaling learning algorithms towards {AI}.
\newblock In \emph{Large Scale Kernel Machines}. MIT Press.

\bibitem[{Berner and Graber(2008)}]{Berner2008OverconfidenceAA}
Eta~S. Berner and Mark~L Graber. 2008.
\newblock \href {https://api.semanticscholar.org/CorpusID:1659486} {Overconfidence as a cause of diagnostic error in medicine.}
\newblock \emph{The American journal of medicine}, 121 5 Suppl:S2--23.

\bibitem[{Cheng et~al.(2023)Cheng, Durmus, and Jurafsky}]{Cheng2023MarkedPU}
Myra Cheng, Esin Durmus, and Dan Jurafsky. 2023.
\newblock \href {https://api.semanticscholar.org/CorpusID:258960243} {Marked personas: Using natural language prompts to measure stereotypes in language models}.
\newblock \emph{ArXiv}, abs/2305.18189.

\bibitem[{Cuadron et~al.(2025)Cuadron, Li, Ma, Wang, Wang, Zhuang, Liu, Schroeder, Xia, Mao et~al.}]{cuadron2025danger}
Alejandro Cuadron, Dacheng Li, Wenjie Ma, Xingyao Wang, Yichuan Wang, Siyuan Zhuang, Shu Liu, Luis~Gaspar Schroeder, Tian Xia, Huanzhi Mao, et~al. 2025.
\newblock The danger of overthinking: Examining the reasoning-action dilemma in agentic tasks.
\newblock \emph{arXiv preprint arXiv:2502.08235}.

\bibitem[{Deroy et~al.(2024)Deroy, Ghosh, and Ghosh}]{Deroy2024ApplicabilityOL}
Aniket Deroy, Kripabandhu Ghosh, and Saptarshi Ghosh. 2024.
\newblock \href {https://api.semanticscholar.org/CorpusID:271222741} {Applicability of large language models and generative models for legal case judgement summarization}.
\newblock \emph{ArXiv}, abs/2407.12848.

\bibitem[{Dong et~al.(2024)Dong, Zhunis, Chin, Han, and Cha}]{dong2024not}
Wenchao Dong, Assem Zhunis, Hyojin Chin, Jiyoung Han, and Meeyoung Cha. 2024.
\newblock I am not them: Fluid identities and persistent out-group bias in large language models.
\newblock \emph{arXiv preprint arXiv:2402.10436}.

\bibitem[{Duan et~al.(2024)Duan, Cheng, Wang, Zavalny, Wang, Xu, Kailkhura, and Xu}]{duan-etal-2024-shifting}
Jinhao Duan, Hao Cheng, Shiqi Wang, Alex Zavalny, Chenan Wang, Renjing Xu, Bhavya Kailkhura, and Kaidi Xu. 2024.
\newblock \href {https://aclanthology.org/2024.acl-long.276} {Shifting attention to relevance: Towards the predictive uncertainty quantification of free-form large language models}.
\newblock In \emph{Proceedings of the 62nd Annual Meeting of the Association for Computational Linguistics (Volume 1: Long Papers)}, pages 5050--5063, Bangkok, Thailand. Association for Computational Linguistics.

\bibitem[{Dubey et~al.(2024)Dubey, Jauhri, Pandey, Kadian, Al-Dahle, Letman, Mathur, Schelten, Yang, Fan, Goyal, Hartshorn, Yang, Mitra, Sravankumar, Korenev, Hinsvark, Rao, Zhang, Rodriguez, Gregerson, Spataru, Rozi{\`e}re, Biron, Tang, Chern, Caucheteux, Nayak, Bi, Marra, McConnell, Keller, Touret, Wu, Wong, Ferrer, Nikolaidis, Allonsius, Song, Pintz, Livshits, Esiobu, Choudhary, Mahajan, Garcia-Olano, Perino, Hupkes, Lakomkin, AlBadawy, Lobanova, Dinan, Smith, Radenovic, Zhang, Synnaeve, Lee, Anderson, Nail, Mialon, Pang, Cucurell, Nguyen, Korevaar, Xu, Touvron, Zarov, Ibarra, Kloumann, Misra, Evtimov, Copet, Lee, Geffert, Vranes, Park, Mahadeokar, Shah, van~der Linde, Billock, Hong, Lee, Fu, Chi, Huang, Liu, Wang, Yu, Bitton, Spisak, Park, Rocca, Johnstun, Saxe, Jia, Alwala, Upasani, Plawiak, Li, neth Heafield, Stone, El-Arini, Iyer, Malik, Chiu, Bhalla, Rantala-Yeary, van~der Maaten, Chen, Tan, Jenkins, Martin, Madaan, Malo, Blecher, Landzaat, de~Oliveira, Muzzi, Pasupuleti, Singh, Paluri, Kardas,
  Oldham, Rita, Pavlova, Kambadur, Lewis, Si, Singh, Hassan, Goyal, Torabi, Bashlykov, Bogoychev, Chatterji, Duchenne, cCelebi, Alrassy, Zhang, Li, Vasic, Weng, Bhargava, Dubal, Krishnan, Koura, Xu, He, Dong, Srinivasan, Ganapathy, Calderer, Cabral, Stojnic, Raileanu, Girdhar, Patel, Sauvestre, Polidoro, Sumbaly, Taylor, Silva, Hou, Wang, Hosseini, Chennabasappa, Singh, Bell, Kim, Edunov, Nie, Narang, Raparthy, Shen, Wan, Bhosale, Zhang, Vandenhende, Batra, Whitman, Sootla, Collot, Gururangan, Borodinsky, Herman, Fowler, Sheasha, Georgiou, Scialom, Speckbacher, Mihaylov, Xiao, Karn, Goswami, Gupta, Ramanathan, Kerkez, Gonguet, Do, Vogeti, Petrovic, Chu, Xiong, Fu, Meers, Martinet, Wang, Tan, Xie, Jia, Wang, Goldschlag, Gaur, Babaei, Wen, Song, Zhang, Li, Mao, Coudert, Yan, Chen, Papakipos, Singh, Grattafiori, Jain, Kelsey, Shajnfeld, Gangidi, Victoria, Goldstand, Menon, Sharma, Boesenberg, Vaughan, Baevski, Feinstein, Kallet, Sangani, Yunus, Lupu, Alvarado, Caples, Gu, Ho, Poulton, Ryan, Ramchandani, Franco,
  Saraf, Chowdhury, Gabriel, Bharambe, Eisenman, Yazdan, James, Maurer, Leonhardi, Huang, Loyd, Paola, Paranjape, Liu, Wu, Ni, Hancock, Wasti, Spence, Stojkovic, Gamido, Montalvo, Parker, Burton, Mejia, Wang, Kim, Zhou, Hu, Chu, Cai, Tindal, Feichtenhofer, Civin, Beaty, Kreymer, Li, Wyatt, Adkins, Xu, Testuggine, David, Parikh, Liskovich, Foss, Wang, Le, Holland, Dowling, Jamil, Montgomery, Presani, Hahn, Wood, Brinkman, Arcaute, Dunbar, Smothers, Sun, Kreuk, Tian, Ozgenel, Caggioni, Guzm'an, Kanayet, Seide, Florez, Schwarz, Badeer, Swee, Halpern, Thattai, Herman, Sizov, Zhang, Lakshminarayanan, Shojanazeri, Zou, Wang, Zha, Habeeb, Rudolph, Suk, Aspegren, Goldman, Molybog, Tufanov, Veliche, Gat, Weissman, Geboski, Kohli, Asher, Gaya, Marcus, Tang, Chan, Zhen, Reizenstein, Teboul, Zhong, Jin, Yang, Cummings, Carvill, Shepard, McPhie, Torres, Ginsburg, Wang, Wu, KamHou, Saxena, Prasad, Khandelwal, Zand, Matosich, Veeraraghavan, Michelena, Li, Huang, Chawla, Lakhotia, Huang, Chen, Garg, Lavender, Silva, Bell,
  Zhang, Guo, Yu, Moshkovich, Wehrstedt, Khabsa, Avalani, Bhatt, Tsimpoukelli, Mankus, Hasson, Lennie, Reso, Groshev, Naumov, Lathi, Keneally, Seltzer, Valko, Restrepo, Patel, Vyatskov, Samvelyan, Clark, Macey, Wang, Hermoso, Metanat, Rastegari, Bansal, Santhanam, Parks, White, Bawa, Singhal, Egebo, Usunier, Laptev, Dong, Zhang, Cheng, Chernoguz, Hart, Salpekar, Kalinli, Kent, Parekh, Saab, Balaji, Rittner, Bontrager, Roux, Doll{\'a}r, Zvyagina, Ratanchandani, Yuvraj, Liang, Alao, Rodriguez, Ayub, Murthy, Nayani, Mitra, Li, Hogan, Battey, Wang, Maheswari, Howes, Rinott, Bondu, Datta, Chugh, Hunt, Dhillon, Sidorov, Pan, Verma, Yamamoto, Ramaswamy, Lindsay, Feng, Lin, Zha, Shankar, Zhang, Wang, Agarwal, Sajuyigbe, Chintala, Max, Chen, Kehoe, Satterfield, Govindaprasad, Gupta, Cho, Virk, Subramanian, Choudhury, Goldman, Remez, Glaser, Best, Kohler, Robinson, Li, Zhang, Matthews, Chou, Shaked, Vontimitta, Ajayi, Montanez, Mohan, Kumar, Mangla, Ionescu, Poenaru, Mihailescu, Ivanov, Li, Wang, Jiang, Bouaziz,
  Constable, Tang, Wang, Wu, Wang, Xia, Wu, Gao, Chen, Hu, Jia, Qi, Li, Zhang, Zhang, Adi, Nam, Wang, Hao, Qian, He, Rait, DeVito, Rosnbrick, Wen, Yang, and Zhao}]{Dubey2024TheL3}
Abhimanyu Dubey, Abhinav Jauhri, Abhinav Pandey, Abhishek Kadian, Ahmad Al-Dahle, Aiesha Letman, Akhil Mathur, Alan Schelten, Amy Yang, Angela Fan, Anirudh Goyal, Anthony Hartshorn, Aobo Yang, Archi Mitra, Archie Sravankumar, Artem Korenev, Arthur Hinsvark, Arun Rao, Aston Zhang, Aurelien Rodriguez, Austen Gregerson, Ava Spataru, Baptiste Rozi{\`e}re, Bethany Biron, Binh Tang, Bobbie Chern, Charlotte Caucheteux, Chaya Nayak, Chloe Bi, Chris Marra, Chris McConnell, Christian Keller, Christophe Touret, Chunyang Wu, Corinne Wong, Cristian~Cant{\'o}n Ferrer, Cyrus Nikolaidis, Damien Allonsius, Daniel Song, Danielle Pintz, Danny Livshits, David Esiobu, Dhruv Choudhary, Dhruv Mahajan, Diego Garcia-Olano, Diego Perino, Dieuwke Hupkes, Egor Lakomkin, Ehab~A. AlBadawy, Elina Lobanova, Emily Dinan, Eric~Michael Smith, Filip Radenovic, Frank Zhang, Gabriele Synnaeve, Gabrielle Lee, Georgia~Lewis Anderson, Graeme Nail, Gr{\'e}goire Mialon, Guanglong Pang, Guillem Cucurell, Hailey Nguyen, Hannah Korevaar, Hu~Xu, Hugo
  Touvron, Iliyan Zarov, Imanol~Arrieta Ibarra, Isabel~M. Kloumann, Ishan Misra, Ivan Evtimov, Jade Copet, Jaewon Lee, Jan~Laurens Geffert, Jana Vranes, Jason Park, Jay Mahadeokar, Jeet Shah, Jelmer van~der Linde, Jennifer Billock, Jenny Hong, Jenya Lee, Jeremy Fu, Jianfeng Chi, Jianyu Huang, Jiawen Liu, Jie Wang, Jiecao Yu, Joanna Bitton, Joe Spisak, Jongsoo Park, Joseph Rocca, Joshua Johnstun, Joshua Saxe, Ju-Qing Jia, Kalyan~Vasuden Alwala, K.~Upasani, Kate Plawiak, Keqian Li, Ken-591 neth Heafield, Kevin Stone, Khalid El-Arini, Krithika Iyer, Kshitiz Malik, Kuenley Chiu, Kunal Bhalla, Lauren Rantala-Yeary, Laurens van~der Maaten, Lawrence Chen, Liang Tan, Liz Jenkins, Louis Martin, Lovish Madaan, Lubo Malo, Lukas Blecher, Lukas Landzaat, Luke de~Oliveira, Madeline~C. Muzzi, Mahesh~Babu Pasupuleti, Mannat Singh, Manohar Paluri, Marcin Kardas, Mathew Oldham, Mathieu Rita, Maya Pavlova, Melissa Hall~Melanie Kambadur, Mike Lewis, Min Si, Mitesh~Kumar Singh, Mona Hassan, Naman Goyal, Narjes Torabi, Nikolay
  Bashlykov, Nikolay Bogoychev, Niladri~S. Chatterji, Olivier Duchenne, Onur cCelebi, Patrick Alrassy, Pengchuan Zhang, Pengwei Li, Petar Vasic, Peter Weng, Prajjwal Bhargava, Pratik Dubal, Praveen Krishnan, Punit~Singh Koura, Puxin Xu, Qing He, Qingxiao Dong, Ragavan Srinivasan, Raj Ganapathy, Ramon Calderer, Ricardo~Silveira Cabral, Robert Stojnic, Roberta Raileanu, Rohit Girdhar, Rohit Patel, Romain Sauvestre, Ronnie Polidoro, Roshan Sumbaly, Ross Taylor, Ruan Silva, Rui Hou, Rui Wang, Saghar Hosseini, Sahana Chennabasappa, Sanjay Singh, Sean Bell, Seohyun~Sonia Kim, Sergey Edunov, Shaoliang Nie, Sharan Narang, Sharath~Chandra Raparthy, Sheng Shen, Shengye Wan, Shruti Bhosale, Shun Zhang, Simon Vandenhende, Soumya Batra, Spencer Whitman, Sten Sootla, Stephane Collot, Suchin Gururangan, Sydney Borodinsky, Tamar Herman, Tara Fowler, Tarek Sheasha, Thomas Georgiou, Thomas Scialom, Tobias Speckbacher, Todor Mihaylov, Tong Xiao, Ujjwal Karn, Vedanuj Goswami, Vibhor Gupta, Vignesh Ramanathan, Viktor Kerkez,
  Vincent Gonguet, Virginie Do, Vish Vogeti, Vladan Petrovic, Weiwei Chu, Wenhan Xiong, Wenyin Fu, Whitney Meers, Xavier Martinet, Xiaodong Wang, Xiaoqing~Ellen Tan, Xinfeng Xie, Xuchao Jia, Xuewei Wang, Yaelle Goldschlag, Yashesh Gaur, Yasmine Babaei, Yiqian Wen, Yiwen Song, Yuchen Zhang, Yue Li, Yuning Mao, Zacharie~Delpierre Coudert, Zhengxu Yan, Zhengxing Chen, Zoe Papakipos, Aaditya~K. Singh, Aaron Grattafiori, Abha Jain, Adam Kelsey, Adam Shajnfeld, Adi Gangidi, Adolfo Victoria, Ahuva Goldstand, Ajay Menon, Ajay Sharma, Alex Boesenberg, Alex Vaughan, Alexei Baevski, Allie Feinstein, Amanda Kallet, Amit Sangani, Anam Yunus, Andrei Lupu, Andres Alvarado, Andrew Caples, Andrew Gu, Andrew Ho, Andrew Poulton, Andrew Ryan, Ankit Ramchandani, Annie Franco, Aparajita Saraf, Arkabandhu Chowdhury, Ashley Gabriel, Ashwin Bharambe, Assaf Eisenman, Azadeh Yazdan, Beau James, Ben Maurer, Ben Leonhardi, Bernie Huang, Beth Loyd, Beto~De Paola, Bhargavi Paranjape, Bing Liu, Bo~Wu, Boyu Ni, Braden Hancock, Bram Wasti,
  Brandon Spence, Brani Stojkovic, Brian Gamido, Britt Montalvo, Carl Parker, Carly Burton, Catalina Mejia, Changhan Wang, Changkyu Kim, Chao Zhou, Chester Hu, Ching-Hsiang Chu, Chris Cai, Chris Tindal, Christoph Feichtenhofer, Damon Civin, Dana Beaty, Daniel Kreymer, Shang-Wen Li, Danny Wyatt, David Adkins, David Xu, Davide Testuggine, Delia David, Devi Parikh, Diana Liskovich, Didem Foss, Dingkang Wang, Duc Le, Dustin Holland, Edward Dowling, Eissa Jamil, Elaine Montgomery, Eleonora Presani, Emily Hahn, Emily Wood, Erik Brinkman, Esteban Arcaute, Evan Dunbar, Evan Smothers, Fei Sun, Felix Kreuk, Feng Tian, Firat Ozgenel, Francesco Caggioni, Francisco Guzm'an, Frank~J. Kanayet, Frank Seide, Gabriela~Medina Florez, Gabriella Schwarz, Gada Badeer, Georgia Swee, Gil Halpern, Govind Thattai, Grant Herman, Grigory~G. Sizov, Guangyi Zhang, Guna Lakshminarayanan, Hamid Shojanazeri, Han Zou, Hannah Wang, Han Zha, Haroun Habeeb, Harrison Rudolph, Helen Suk, Henry Aspegren, Hunter Goldman, Igor Molybog, Igor Tufanov,
  Irina-Elena Veliche, Itai Gat, Jake Weissman, James Geboski, James Kohli, Japhet Asher, Jean-Baptiste Gaya, Jeff Marcus, Jeff Tang, Jennifer Chan, Jenny Zhen, Jeremy Reizenstein, Jeremy Teboul, Jessica Zhong, Jian Jin, Jingyi Yang, Joe Cummings, Jon Carvill, Jon Shepard, Jonathan McPhie, Jonathan Torres, Josh Ginsburg, Junjie Wang, Kaixing(Kai) Wu, U~KamHou, Karan Saxena, Karthik Prasad, Kartikay Khandelwal, Katayoun Zand, Kathy Matosich, Kaushik Veeraraghavan, Kelly Michelena, Keqian Li, Kun Huang, Kunal Chawla, Kushal Lakhotia, Kyle Huang, Lailin Chen, Lakshya Garg, A~Lavender, Leandro Silva, Lee Bell, Lei Zhang, Liangpeng Guo, Licheng Yu, Liron Moshkovich, Luca Wehrstedt, Madian Khabsa, Manav Avalani, Manish Bhatt, Maria Tsimpoukelli, Martynas Mankus, Matan Hasson, Matthew Lennie, Matthias Reso, Maxim Groshev, Maxim Naumov, Maya Lathi, Meghan Keneally, Michael~L. Seltzer, Michal Valko, Michelle Restrepo, Mihir Patel, Mik Vyatskov, Mikayel Samvelyan, Mike Clark, Mike Macey, Mike Wang, Miquel~Jubert
  Hermoso, Mo~Metanat, Mohammad Rastegari, Munish Bansal, Nandhini Santhanam, Natascha Parks, Natasha White, Navyata Bawa, Nayan Singhal, Nick Egebo, Nicolas Usunier, Nikolay~Pavlovich Laptev, Ning Dong, Ning Zhang, Norman Cheng, Oleg Chernoguz, Olivia Hart, Omkar Salpekar, Ozlem Kalinli, Parkin Kent, Parth Parekh, Paul Saab, Pavan Balaji, Pedro Rittner, Philip Bontrager, Pierre Roux, Piotr Doll{\'a}r, Polina Zvyagina, Prashant Ratanchandani, Pritish Yuvraj, Qian Liang, Rachad Alao, Rachel Rodriguez, Rafi Ayub, Raghotham Murthy, Raghu Nayani, Rahul Mitra, Raymond Li, Rebekkah Hogan, Robin Battey, Rocky Wang, Rohan Maheswari, Russ Howes, Ruty Rinott, Sai~Jayesh Bondu, Samyak Datta, Sara Chugh, Sara Hunt, Sargun Dhillon, Sasha Sidorov, Satadru Pan, Saurabh Verma, Seiji Yamamoto, Sharadh Ramaswamy, Shaun Lindsay, Sheng Feng, Shenghao Lin, Shengxin~Cindy Zha, Shiva Shankar, Shuqiang Zhang, Sinong Wang, Sneha Agarwal, Soji Sajuyigbe, Soumith Chintala, Stephanie Max, Stephen Chen, Steve Kehoe, Steve Satterfield,
  Sudarshan Govindaprasad, Sumit Gupta, Sung-Bae Cho, Sunny Virk, Suraj Subramanian, Sy~Choudhury, Sydney Goldman, Tal Remez, Tamar Glaser, Tamara Best, Thilo Kohler, Thomas Robinson, Tianhe Li, Tianjun Zhang, Tim Matthews, Timothy Chou, Tzook Shaked, Varun Vontimitta, Victoria Ajayi, Victoria Montanez, Vijai Mohan, Vinay~Satish Kumar, Vishal Mangla, Vlad Ionescu, Vlad~Andrei Poenaru, Vlad~T. Mihailescu, Vladimir Ivanov, Wei Li, Wenchen Wang, Wenwen Jiang, Wes Bouaziz, Will Constable, Xia Tang, Xiaofang Wang, Xiaojian Wu, Xiaolan Wang, Xide Xia, Xilun Wu, Xinbo Gao, Yanjun Chen, Ye~Hu, Ye~Jia, Ye~Qi, Yenda Li, Yilin Zhang, Ying Zhang, Yossi Adi, Youngjin Nam, Yu~Wang, Yuchen Hao, Yundi Qian, Yuzi He, Zach Rait, Zachary DeVito, Zef Rosnbrick, Zhaoduo Wen, Zhenyu Yang, and Zhiwei Zhao. 2024.
\newblock \href {https://api.semanticscholar.org/CorpusID:271571434} {The llama 3 herd of models}.
\newblock \emph{ArXiv}, abs/2407.21783.

\bibitem[{Feng et~al.(2023)Feng, Park, Liu, and Tsvetkov}]{feng-etal-2023-pretraining}
Shangbin Feng, Chan~Young Park, Yuhan Liu, and Yulia Tsvetkov. 2023.
\newblock \href {https://doi.org/10.18653/v1/2023.acl-long.656} {From pretraining data to language models to downstream tasks: Tracking the trails of political biases leading to unfair {NLP} models}.
\newblock In \emph{Proceedings of the 61st Annual Meeting of the Association for Computational Linguistics (Volume 1: Long Papers)}, pages 11737--11762, Toronto, Canada. Association for Computational Linguistics.

\bibitem[{Gekhman et~al.(2024)Gekhman, Yona, Aharoni, Eyal, Feder, Reichart, and Herzig}]{Gekhman2024DoesFL}
Zorik Gekhman, G.~Yona, Roee Aharoni, Matan Eyal, Amir Feder, Roi Reichart, and Jonathan Herzig. 2024.
\newblock \href {https://api.semanticscholar.org/CorpusID:269635770} {Does fine-tuning llms on new knowledge encourage hallucinations?}
\newblock \emph{ArXiv}, abs/2405.05904.

\bibitem[{Goodfellow et~al.(2016)Goodfellow, Bengio, Courville, and Bengio}]{goodfellow2016deep}
Ian Goodfellow, Yoshua Bengio, Aaron Courville, and Yoshua Bengio. 2016.
\newblock \emph{Deep learning}, volume~1.
\newblock MIT Press.

\bibitem[{Guo et~al.(2017{\natexlab{a}})Guo, Pleiss, Sun, and Weinberger}]{guo2017calibration}
Chuan Guo, Geoff Pleiss, Yu~Sun, and Kilian~Q Weinberger. 2017{\natexlab{a}}.
\newblock On calibration of modern neural networks.
\newblock In \emph{International conference on machine learning}, pages 1321--1330. PMLR.

\bibitem[{Guo et~al.(2017{\natexlab{b}})Guo, Pleiss, Sun, and Weinberger}]{Guo2017OnCO}
Chuan Guo, Geoff Pleiss, Yu~Sun, and Kilian~Q. Weinberger. 2017{\natexlab{b}}.
\newblock \href {http://proceedings.mlr.press/v70/guo17a.html} {On calibration of modern neural networks}.
\newblock In \emph{Proceedings of the 34th International Conference on Machine Learning, {ICML} 2017, Sydney, NSW, Australia, 6-11 August 2017}, volume~70 of \emph{Proceedings of Machine Learning Research}, pages 1321--1330. {PMLR}.

\bibitem[{Hada et~al.(2023)Hada, Seth, Diddee, and Bali}]{hada-etal-2023-fifty}
Rishav Hada, Agrima Seth, Harshita Diddee, and Kalika Bali. 2023.
\newblock \href {https://doi.org/10.18653/v1/2023.emnlp-main.115} {{``}fifty shades of bias{''}: Normative ratings of gender bias in {GPT} generated {E}nglish text}.
\newblock In \emph{Proceedings of the 2023 Conference on Empirical Methods in Natural Language Processing}, pages 1862--1876, Singapore. Association for Computational Linguistics.

\bibitem[{Hagendorff(2023)}]{hagendorff2023machine}
Thilo Hagendorff. 2023.
\newblock Machine psychology: Investigating emergent capabilities and behavior in large language models using psychological methods.
\newblock \emph{arXiv preprint arXiv:2303.13988}, 1.

\bibitem[{Hase et~al.(2024{\natexlab{a}})Hase, Bansal, Clark, and Wiegreffe}]{hase2024unreasonable}
Peter Hase, Mohit Bansal, Peter Clark, and Sarah Wiegreffe. 2024{\natexlab{a}}.
\newblock The unreasonable effectiveness of easy training data for hard tasks.
\newblock \emph{arXiv preprint arXiv:2401.06751}.

\bibitem[{Hase et~al.(2024{\natexlab{b}})Hase, Bansal, Clark, and Wiegreffe}]{Hase2024TheUE}
Peter Hase, Mohit Bansal, Peter Clark, and Sarah Wiegreffe. 2024{\natexlab{b}}.
\newblock \href {https://api.semanticscholar.org/CorpusID:266977266} {The unreasonable effectiveness of easy training data for hard tasks}.
\newblock \emph{ArXiv}, abs/2401.06751.

\bibitem[{Hendrycks et~al.(2021)Hendrycks, Burns, Basart, Zou, Mazeika, Song, and Steinhardt}]{hendrycks2021measuring}
Dan Hendrycks, Collin Burns, Steven Basart, Andy Zou, Mantas Mazeika, Dawn Song, and Jacob Steinhardt. 2021.
\newblock \href {https://openreview.net/forum?id=d7KBjmI3GmQ} {Measuring massive multitask language understanding}.
\newblock In \emph{International Conference on Learning Representations}.

\bibitem[{Hinton et~al.(2006)Hinton, Osindero, and Teh}]{Hinton06}
Geoffrey~E. Hinton, Simon Osindero, and Yee~Whye Teh. 2006.
\newblock A fast learning algorithm for deep belief nets.
\newblock \emph{Neural Computation}, 18:1527--1554.

\bibitem[{Jiang et~al.(2023)Jiang, Zhang, Cao, Breazeal, Roy, and Kabbara}]{jiang2023personallm}
Hang Jiang, Xiajie Zhang, Xubo Cao, Cynthia Breazeal, Deb Roy, and Jad Kabbara. 2023.
\newblock Personallm: Investigating the ability of large language models to express personality traits.
\newblock \emph{arXiv preprint arXiv:2305.02547}.

\bibitem[{Kadavath et~al.(2022)Kadavath, Conerly, Askell, Henighan, Drain, Perez, Schiefer, Dodds, DasSarma, Tran-Johnson et~al.}]{kadavath2022language}
Saurav Kadavath, Tom Conerly, Amanda Askell, Tom Henighan, Dawn Drain, Ethan Perez, Nicholas Schiefer, Zac~Hatfield Dodds, Nova DasSarma, Eli Tran-Johnson, et~al. 2022.
\newblock \href {https://arxiv.org/abs/2207.05221} {Language models (mostly) know what they know}.
\newblock \emph{ArXiv preprint}, abs/2207.05221.

\bibitem[{Kim et~al.(2024)Kim, Liao, Vorvoreanu, Ballard, and Vaughan}]{Kim2024ImNS}
Sunnie S.~Y. Kim, Q.~Vera Liao, Mihaela Vorvoreanu, Steph Ballard, and Jennifer~Wortman Vaughan. 2024.
\newblock \href {https://api.semanticscholar.org/CorpusID:269484145} {"i'm not sure, but...": Examining the impact of large language models' uncertainty expression on user reliance and trust}.
\newblock \emph{Proceedings of the 2024 ACM Conference on Fairness, Accountability, and Transparency}.

\bibitem[{Kruger and Dunning(1999)}]{kruger1999unskilled}
Justin Kruger and David Dunning. 1999.
\newblock Unskilled and unaware of it: how difficulties in recognizing one's own incompetence lead to inflated self-assessments.
\newblock \emph{Journal of personality and social psychology}, 77(6):1121.

\bibitem[{Kuhn et~al.(2023)Kuhn, Gal, and Farquhar}]{kuhn2023semantic}
Lorenz Kuhn, Yarin Gal, and Sebastian Farquhar. 2023.
\newblock \href {https://arxiv.org/abs/2302.09664} {Semantic uncertainty: Linguistic invariances for uncertainty estimation in natural language generation}.
\newblock \emph{ArXiv preprint}, abs/2302.09664.

\bibitem[{Kumar et~al.(2024)Kumar, Morabito, Umbet, Kabbara, and Emami}]{kumar-etal-2024-confidence}
Abhishek Kumar, Robert Morabito, Sanzhar Umbet, Jad Kabbara, and Ali Emami. 2024.
\newblock \href {https://doi.org/10.18653/v1/2024.acl-long.20} {Confidence under the hood: An investigation into the confidence-probability alignment in large language models}.
\newblock In \emph{Proceedings of the 62nd Annual Meeting of the Association for Computational Linguistics (Volume 1: Long Papers)}, pages 315--334, Bangkok, Thailand. Association for Computational Linguistics.

\bibitem[{Leng et~al.(2025)Leng, Huang, Zhu, and Huang}]{leng2025tamingoverconfidencellmsreward}
Jixuan Leng, Chengsong Huang, Banghua Zhu, and Jiaxin Huang. 2025.
\newblock \href {https://arxiv.org/abs/2410.09724} {Taming overconfidence in llms: Reward calibration in rlhf}.
\newblock \emph{Preprint}, arXiv:2410.09724.

\bibitem[{Li(2023)}]{Li2023TheDS}
Zihao~(Michael) Li. 2023.
\newblock \href {https://api.semanticscholar.org/CorpusID:258352281} {The dark side of chatgpt: Legal and ethical challenges from stochastic parrots and hallucination}.
\newblock \emph{ArXiv}, abs/2304.14347.

\bibitem[{Light et~al.(2022)Light, Fernbach, Rabb, Geana, and Sloman}]{light2022knowledge}
Nicholas Light, Philip~M Fernbach, Nathaniel Rabb, Mugur~V Geana, and Steven~A Sloman. 2022.
\newblock Knowledge overconfidence is associated with anti-consensus views on controversial scientific issues.
\newblock \emph{Science Advances}, 8(29):eabo0038.

\bibitem[{Lin et~al.(2022)Lin, Hilton, and Evans}]{lin2022teaching}
Stephanie Lin, Jacob Hilton, and Owain Evans. 2022.
\newblock \href {https://openreview.net/forum?id=8s8K2UZGTZ} {Teaching models to express their uncertainty in words}.
\newblock \emph{Transactions on Machine Learning Research}.

\bibitem[{Lin et~al.(2023)Lin, Trivedi, and Sun}]{lin2023generating}
Zhen Lin, Shubhendu Trivedi, and Jimeng Sun. 2023.
\newblock \href {https://arxiv.org/abs/2305.19187} {Generating with confidence: Uncertainty quantification for black-box large language models}.
\newblock \emph{ArXiv preprint}, abs/2305.19187.

\bibitem[{Mbakwe et~al.(2023)Mbakwe, Lourentzou, Celi, Mechanic, and Dagan}]{10.1371/journal.pdig.0000205}
Amarachi~B. Mbakwe, Ismini Lourentzou, Leo~Anthony Celi, Oren~J. Mechanic, and Alon Dagan. 2023.
\newblock \href {https://doi.org/10.1371/journal.pdig.0000205} {Chatgpt passing usmle shines a spotlight on the flaws of medical education}.
\newblock \emph{PLOS Digital Health}, 2(2):1--3.

\bibitem[{Mielke et~al.(2022)Mielke, Szlam, Dinan, and Boureau}]{mielke2022reducing}
Sabrina~J. Mielke, Arthur Szlam, Emily Dinan, and Y-Lan Boureau. 2022.
\newblock \href {https://doi.org/10.1162/tacl_a_00494} {Reducing conversational agents{'} overconfidence through linguistic calibration}.
\newblock \emph{Transactions of the Association for Computational Linguistics}, 10:857--872.

\bibitem[{Moore and Healy(2008)}]{moore2008trouble}
Don~A Moore and Paul~J Healy. 2008.
\newblock The trouble with overconfidence.
\newblock \emph{Psychological review}, 115(2):502.

\bibitem[{Park et~al.(2023)Park, O'Brien, Cai, Morris, Liang, and Bernstein}]{10.1145/3586183.3606763}
Joon~Sung Park, Joseph O'Brien, Carrie~Jun Cai, Meredith~Ringel Morris, Percy Liang, and Michael~S. Bernstein. 2023.
\newblock \href {https://doi.org/10.1145/3586183.3606763} {Generative agents: Interactive simulacra of human behavior}.
\newblock In \emph{Proceedings of the 36th Annual ACM Symposium on User Interface Software and Technology}, UIST '23, New York, NY, USA. Association for Computing Machinery.

\bibitem[{Park et~al.(2022)Park, Popowski, Cai, Morris, Liang, and Bernstein}]{park2022social}
Joon~Sung Park, Lindsay Popowski, Carrie Cai, Meredith~Ringel Morris, Percy Liang, and Michael~S Bernstein. 2022.
\newblock Social simulacra: Creating populated prototypes for social computing systems.
\newblock In \emph{Proceedings of the 35th Annual ACM Symposium on User Interface Software and Technology}, pages 1--18.

\bibitem[{Qian et~al.(2023)Qian, Cong, Yang, Chen, Su, Xu, Liu, and Sun}]{qian2023communicative}
Chen Qian, Xin Cong, Cheng Yang, Weize Chen, Yusheng Su, Juyuan Xu, Zhiyuan Liu, and Maosong Sun. 2023.
\newblock Communicative agents for software development.
\newblock \emph{arXiv preprint arXiv:2307.07924}, 6(3).

\bibitem[{Rein et~al.(2023)Rein, Hou, Stickland, Petty, Pang, Dirani, Michael, and Bowman}]{rein2023gpqa}
David Rein, Betty~Li Hou, Asa~Cooper Stickland, Jackson Petty, Richard~Yuanzhe Pang, Julien Dirani, Julian Michael, and Samuel~R Bowman. 2023.
\newblock Gpqa: A graduate-level google-proof q\&a benchmark.
\newblock \emph{arXiv preprint arXiv:2311.12022}.

\bibitem[{R{\'i}os-Hoyo et~al.(2024)R{\'i}os-Hoyo, Shan, Li, Pearson, Pusztai, and Howard}]{RosHoyo2024EvaluationOL}
Alejandro R{\'i}os-Hoyo, Naing~Lin Shan, Anran Li, Alexander~T. Pearson, Lajos Pusztai, and Frederick~M. Howard. 2024.
\newblock \href {https://api.semanticscholar.org/CorpusID:270656792} {Evaluation of large language models as a diagnostic aid for complex medical cases}.
\newblock \emph{Frontiers in Medicine}, 11.

\bibitem[{Sanchez and Dunning(2018)}]{sanchez2018overconfidence}
Carmen Sanchez and David Dunning. 2018.
\newblock Overconfidence among beginners: Is a little learning a dangerous thing?
\newblock \emph{Journal of personality and Social Psychology}, 114(1):10.

\bibitem[{Sanchez and Dunning(2024)}]{sanchez2024intermediate}
Carmen Sanchez and David Dunning. 2024.
\newblock Intermediate science knowledge predicts overconfidence.
\newblock \emph{Trends in Cognitive Sciences}, 28(4):284--285.

\bibitem[{Santurkar et~al.(2023)Santurkar, Durmus, Ladhak, Lee, Liang, and Hashimoto}]{santurkar2023whose}
Shibani Santurkar, Esin Durmus, Faisal Ladhak, Cinoo Lee, Percy Liang, and Tatsunori Hashimoto. 2023.
\newblock Whose opinions do language models reflect?
\newblock In \emph{International Conference on Machine Learning}, pages 29971--30004. PMLR.

\bibitem[{Sap et~al.(2022)Sap, Le~Bras, Fried, and Choi}]{sap-etal-2022-neural}
Maarten Sap, Ronan Le~Bras, Daniel Fried, and Yejin Choi. 2022.
\newblock \href {https://doi.org/10.18653/v1/2022.emnlp-main.248} {Neural theory-of-mind? on the limits of social intelligence in large {LM}s}.
\newblock In \emph{Proceedings of the 2022 Conference on Empirical Methods in Natural Language Processing}, pages 3762--3780, Abu Dhabi, United Arab Emirates. Association for Computational Linguistics.

\bibitem[{Schramowski et~al.(2022)Schramowski, Turan, Andersen, Rothkopf, and Kersting}]{schramowski2022large}
Patrick Schramowski, Cigdem Turan, Nico Andersen, Constantin~A Rothkopf, and Kristian Kersting. 2022.
\newblock Large pre-trained language models contain human-like biases of what is right and wrong to do.
\newblock \emph{Nature Machine Intelligence}, 4(3):258--268.

\bibitem[{Sclar et~al.(2023)Sclar, Choi, Tsvetkov, and Suhr}]{sclar2023quantifying}
Melanie Sclar, Yejin Choi, Yulia Tsvetkov, and Alane Suhr. 2023.
\newblock Quantifying language models' sensitivity to spurious features in prompt design or: How i learned to start worrying about prompt formatting.
\newblock \emph{arXiv preprint arXiv:2310.11324}.

\bibitem[{Shah et~al.(2023)Shah, Pour, Tagade, Casper, Rando et~al.}]{shah2023scalable}
Rusheb Shah, Soroush Pour, Arush Tagade, Stephen Casper, Javier Rando, et~al. 2023.
\newblock Scalable and transferable black-box jailbreaks for language models via persona modulation.
\newblock \emph{arXiv preprint arXiv:2311.03348}.

\bibitem[{Shanahan et~al.(2023)Shanahan, McDonell, and Reynolds}]{shanahan2023role}
Murray Shanahan, Kyle McDonell, and Laria Reynolds. 2023.
\newblock Role play with large language models.
\newblock \emph{Nature}, 623(7987):493--498.

\bibitem[{Teplica et~al.(2025)Teplica, Liu, Cohan, and Rudner}]{teplicasciurus}
Carter Teplica, Yixin Liu, Arman Cohan, and Tim~GJ Rudner. 2025.
\newblock Sciurus: Shared circuits for interpretable uncertainty representations in language models.
\newblock In \emph{MINT: Foundation Model Interventions}.

\bibitem[{Tian et~al.(2023)Tian, Mitchell, Zhou, Sharma, Rafailov, Yao, Finn, and Manning}]{tian-etal-2023-just}
Katherine Tian, Eric Mitchell, Allan Zhou, Archit Sharma, Rafael Rafailov, Huaxiu Yao, Chelsea Finn, and Christopher Manning. 2023.
\newblock \href {https://doi.org/10.18653/v1/2023.emnlp-main.330} {Just ask for calibration: Strategies for eliciting calibrated confidence scores from language models fine-tuned with human feedback}.
\newblock In \emph{Proceedings of the 2023 Conference on Empirical Methods in Natural Language Processing}, pages 5433--5442, Singapore. Association for Computational Linguistics.

\bibitem[{Tjuatja et~al.(2023)Tjuatja, Chen, Wu, Talwalkar, and Neubig}]{tjuatja2023llms}
Lindia Tjuatja, Valerie Chen, Sherry~Tongshuang Wu, Ameet Talwalkar, and Graham Neubig. 2023.
\newblock Do llms exhibit human-like response biases? a case study in survey design.
\newblock \emph{arXiv preprint arXiv:2311.04076}.

\bibitem[{van Prooijen(2021)}]{Prooijen2021OverconfidenceIR}
Jan-Willem van Prooijen. 2021.
\newblock \href {https://api.semanticscholar.org/CorpusID:233884520} {Overconfidence in radical politics}.
\newblock \emph{The Psychology of Populism}.

\bibitem[{Wan et~al.(2023)Wan, Pu, Sun, Garimella, Chang, and Peng}]{wan-etal-2023-kelly}
Yixin Wan, George Pu, Jiao Sun, Aparna Garimella, Kai-Wei Chang, and Nanyun Peng. 2023.
\newblock \href {https://doi.org/10.18653/v1/2023.findings-emnlp.243} {{``}kelly is a warm person, joseph is a role model{''}: Gender biases in {LLM}-generated reference letters}.
\newblock In \emph{Findings of the Association for Computational Linguistics: EMNLP 2023}, pages 3730--3748, Singapore. Association for Computational Linguistics.

\bibitem[{Wang et~al.(2024{\natexlab{a}})Wang, Ma, Hu, Weber-Genzel, R{\"o}ttger, Kreuter, Hovy, and Plank}]{wang2024my}
Xinpeng Wang, Bolei Ma, Chengzhi Hu, Leon Weber-Genzel, Paul R{\"o}ttger, Frauke Kreuter, Dirk Hovy, and Barbara Plank. 2024{\natexlab{a}}.
\newblock " my answer is c": First-token probabilities do not match text answers in instruction-tuned language models.
\newblock \emph{arXiv preprint arXiv:2402.14499}.

\bibitem[{Wang et~al.(2024{\natexlab{b}})Wang, Xiao, Huang, Yuan, Xu, Guo, Tu, Fei, Leng, Wang, Chen, Li, and Xiao}]{wang-etal-2024-incharacter}
Xintao Wang, Yunze Xiao, Jen-tse Huang, Siyu Yuan, Rui Xu, Haoran Guo, Quan Tu, Yaying Fei, Ziang Leng, Wei Wang, Jiangjie Chen, Cheng Li, and Yanghua Xiao. 2024{\natexlab{b}}.
\newblock \href {https://doi.org/10.18653/v1/2024.acl-long.102} {{I}n{C}haracter: Evaluating personality fidelity in role-playing agents through psychological interviews}.
\newblock In \emph{Proceedings of the 62nd Annual Meeting of the Association for Computational Linguistics (Volume 1: Long Papers)}, pages 1840--1873, Bangkok, Thailand. Association for Computational Linguistics.

\bibitem[{Wang et~al.(2023)Wang, Peng, Que, Liu, Zhou, Wu, Guo, Gan, Ni, Yang et~al.}]{wang2023rolellm}
Zekun~Moore Wang, Zhongyuan Peng, Haoran Que, Jiaheng Liu, Wangchunshu Zhou, Yuhan Wu, Hongcheng Guo, Ruitong Gan, Zehao Ni, Jian Yang, et~al. 2023.
\newblock Rolellm: Benchmarking, eliciting, and enhancing role-playing abilities of large language models.
\newblock \emph{arXiv preprint arXiv:2310.00746}.

\bibitem[{Wei et~al.(2024)Wei, Karina, Chung, Jiao, Papay, Glaese, Schulman, and Fedus}]{wei2024measuringshortformfactualitylarge}
Jason Wei, Nguyen Karina, Hyung~Won Chung, Yunxin~Joy Jiao, Spencer Papay, Amelia Glaese, John Schulman, and William Fedus. 2024.
\newblock \href {https://arxiv.org/abs/2411.04368} {Measuring short-form factuality in large language models}.
\newblock \emph{Preprint}, arXiv:2411.04368.

\bibitem[{Wen et~al.(2024{\natexlab{a}})Wen, Howe, and Wang}]{wen-etal-2024-characterizing}
Bingbing Wen, Bill Howe, and Lucy~Lu Wang. 2024{\natexlab{a}}.
\newblock \href {https://doi.org/10.18653/v1/2024.findings-emnlp.197} {Characterizing {LLM} abstention behavior in science {QA} with context perturbations}.
\newblock In \emph{Findings of the Association for Computational Linguistics: EMNLP 2024}, pages 3437--3450, Miami, Florida, USA. Association for Computational Linguistics.

\bibitem[{Wen et~al.(2024{\natexlab{b}})Wen, Xu, HAN, Wolfe, Wang, and Howe}]{wen2024from}
Bingbing Wen, Chenjun Xu, Bin HAN, Robert Wolfe, Lucy~Lu Wang, and Bill Howe. 2024{\natexlab{b}}.
\newblock \href {https://openreview.net/forum?id=y9UdO5cmHs} {From human to model overconfidence: Evaluating confidence dynamics in large language models}.
\newblock In \emph{NeurIPS 2024 Workshop on Behavioral Machine Learning}.

\bibitem[{Wen et~al.(2025)Wen, Yao, Feng, Xu, Tsvetkov, Howe, and Wang}]{wen2024art}
Bingbing Wen, Jihan Yao, Shangbin Feng, Chenjun Xu, Yulia Tsvetkov, Bill Howe, and Lucy~Lu Wang. 2025.
\newblock Know your limits: A survey of abstention in large language models.
\newblock \emph{Transactions of the Association for Computational Linguistics (ACL)}.

\bibitem[{Wong and Kim(2023)}]{wong2023chatgpt}
Jared Wong and Jin Kim. 2023.
\newblock Chatgpt is more likely to be perceived as male than female.
\newblock \emph{arXiv preprint arXiv:2305.12564}.

\bibitem[{Xiong et~al.(2024)Xiong, Hu, Lu, LI, Fu, He, and Hooi}]{xiong2024can}
Miao Xiong, Zhiyuan Hu, Xinyang Lu, YIFEI LI, Jie Fu, Junxian He, and Bryan Hooi. 2024.
\newblock \href {https://openreview.net/forum?id=gjeQKFxFpZ} {Can {LLM}s express their uncertainty? an empirical evaluation of confidence elicitation in {LLM}s}.
\newblock In \emph{The Twelfth International Conference on Learning Representations}.

\bibitem[{Xu et~al.(2024)Xu, Guo, Zhou, Song, and Niu}]{enterprise_decision_support}
Zeqiu Xu, Lingfeng Guo, Shuwen Zhou, Runze Song, and Kaiyi Niu. 2024.
\newblock \href {https://doi.org/10.5281/zenodo.12670581} {Enterprise supply chain risk management and decision support driven by large language models}.
\newblock \emph{Applied Science and Engineering Journal for Advanced Research}, 3(4):1–7.

\bibitem[{Zeng et~al.(2024)Zeng, Jin, Yu, Wang, Hua, Zhou, Sun, Meng, Ma, Wang et~al.}]{zeng2024uncertainty}
Qingcheng Zeng, Mingyu Jin, Qinkai Yu, Zhenting Wang, Wenyue Hua, Zihao Zhou, Guangyan Sun, Yanda Meng, Shiqing Ma, Qifan Wang, et~al. 2024.
\newblock Uncertainty is fragile: Manipulating uncertainty in large language models.
\newblock \emph{arXiv preprint arXiv:2407.11282}.

\bibitem[{Zhou et~al.(2023)Zhou, Jurafsky, and Hashimoto}]{Zhou2023NavigatingTG}
Kaitlyn Zhou, Dan Jurafsky, and Tatsunori Hashimoto. 2023.
\newblock \href {https://api.semanticscholar.org/CorpusID:257220189} {Navigating the grey area: Expressions of overconfidence and uncertainty in language models}.
\newblock \emph{ArXiv}, abs/2302.13439.

\bibitem[{Ziems et~al.(2023)Ziems, Held, Shaikh, Chen, Zhang, and Yang}]{Ziems2023CanLL}
Caleb Ziems, William~B. Held, Omar Shaikh, Jiaao Chen, Zhehao Zhang, and Diyi Yang. 2023.
\newblock \href {https://api.semanticscholar.org/CorpusID:258547324} {Can large language models transform computational social science?}
\newblock \emph{Computational Linguistics}, 50:237--291.

\end{thebibliography}

\newpage

\appendix

\newpage

\section{Appendix / supplemental material}

\subsection{Dataset and Model Details}
\label{tab:dataset_statistics}

This section provides comprehensive details on the datasets and language models. Table~\ref{tab:dataset_statistics} summarizes all the datasets used in this paper, which are drawn from four sources: \textbf{MMLU}, \textbf{GPQA}, \textbf{SimpleQA}, and \textbf{NQ-Open}. Among them, MMLU and GPQA are explicitly structured to include questions of varying difficulty levels, allowing us to analyze how model confidence and performance are affected by task hardness. Table~\ref{tab:model_statistics} lists the specific versions of the models used in this study.

\begin{table}[h!]
    \small
    \centering
    \begin{tabular}{lccc}
        \toprule
        \textbf{Dataset}  & \textbf{Hardness}  & \textbf{Subject} & \textbf{Test Size} \\            
        \midrule
        MMLU   & High School & Physics  & 170 \\
        MMLU   & High School & Chemistry  & 230 \\
        MMLU   & High School & Biology  & 340 \\
        MMLU   & High School & Math  & 300 \\
        MMLU   & High School & Computer Science  & 110 \\
        MMLU   & College & Physics  & 110 \\
        MMLU   & College & Chemistry  & 110 \\
        MMLU   & College & Biology  & 160 \\
        MMLU   & College & Math  & 110 \\
        MMLU   & College & Computer Science  & 110 \\
        MMLU   & College & Medicine  & 200 \\
        MMLU   & Expert & Medicine  & 300 \\
        GPQA   & Expert & Physics  & 220 \\
        GPQA   & Expert & Chemistry  & 210 \\
        GPQA   & Expert & Biology  & 100 \\
        SimpleQA   & general & general  & 100 \\
        NQ-open   & general & general  & 100 \\
    \bottomrule
    \end{tabular}
    \vspace{2pt}
    \caption{Dataset statistics.}
    \label{tab:dataset_statistics}
\end{table}

\begin{table}[h!]
    \scriptsize
    \setlength{\tabcolsep}{4pt}
    \renewcommand{\arraystretch}{0.9}
    \centering
    \begin{tabular}{p{3.5cm}p{2cm}p{1.5cm}}
        \toprule
        \textbf{Model ID}  & \textbf{Date Release}  & \textbf{Developer}  \\            
        \midrule
        gemma2-9b-it   & Jun. 27, 2024 & 	Google  \\
        Meta-Llama-3-70B-Instruct   & Apr. 18, 2024 & Meta \\
        Meta-Llama-3-8B-Instruct   & Apr. 18, 2024 & Meta \\
        Llama-3.2-90B-Vision-Instruct   & Sept. 25, 2024 & Meta \\
        Mixtral-8x7B-Instruct-v0.1    & Dec. 11, 2023  & Mistral \\
        GPT-4o   & May 13, 2024 & OpenAI \\
        claude-3-sonnet-20240229   & Feb. 29, 2024 & Anthropic \\
    \bottomrule
    \end{tabular}
    \vspace{2pt}
    \caption{Model statistics.}
    \label{tab:model_statistics}
\end{table}

\subsection{Overconfidence under Subject and Difficulty}
\label{fig:hardness}

\begin{figure*}[t] 
    \centering
    \includegraphics[width=1\textwidth]{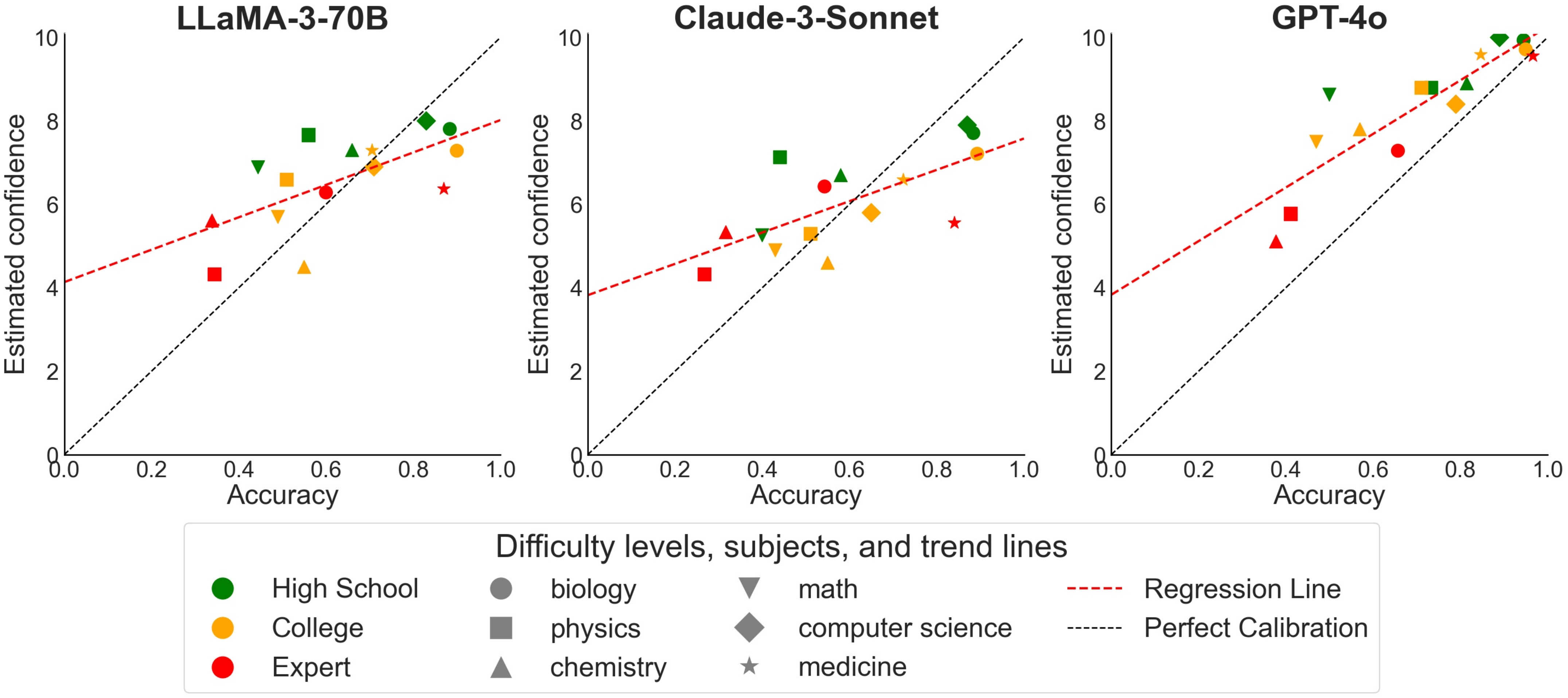} 
    \caption{Confidence estimation using our method on tasks with various levels of difficulty. GPT-4o appears especially more sensitive to task difficulty than LLaMA-3-70B and Claude-3-Sonnet.}
    \label{fig:hardness}
\end{figure*}

Figure~\ref{fig:hardness} shows the performance of three models—\textbf{GPT-4o}, \textbf{LLaMA-3-70B}, and \textbf{Claude-3-Sonnet}—across different subjects and difficulty levels. This visualization highlights how both subject area and task hardness influence model confidence, offering additional insights into the calibration behavior of these large language models. Table~\ref{tab:method_compare_all_appendix} extends Table~\ref{tab:method_compare_all} by including results from four additional models.

\subsection{Overplacement Analysis}
\label{fig:overplacement_combined}

Figure~\ref{fig:overplacement_combined} presents overplacement results across seven models under different role-playing conditions. It visualizes the gap between confidence and accuracy more intuitively for each role. In comparison, the main text's Figure~\ref{fig:overplacement_result_score} focuses on the overall degree of overplacement exhibited by LLMs towards various level of expertise roles, providing a complementary perspective.

\begin{figure*}[h]
    \centering
    \includegraphics[width=1\textwidth]{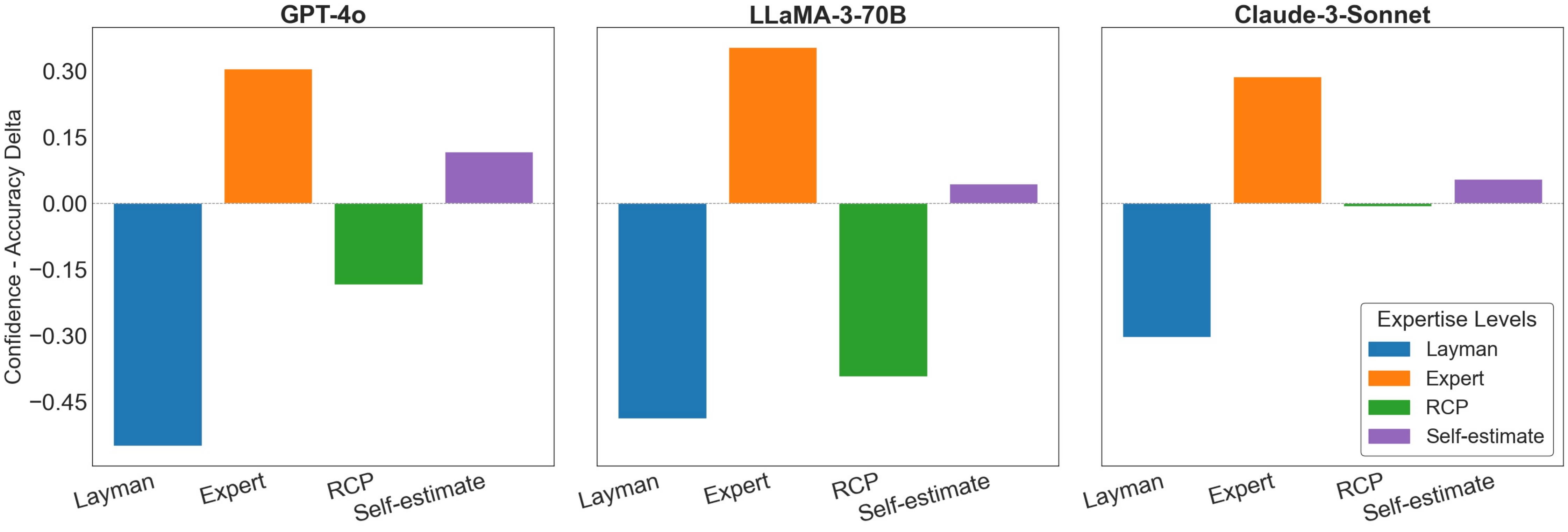} \\[2mm]
    \includegraphics[width=0.9\textwidth]{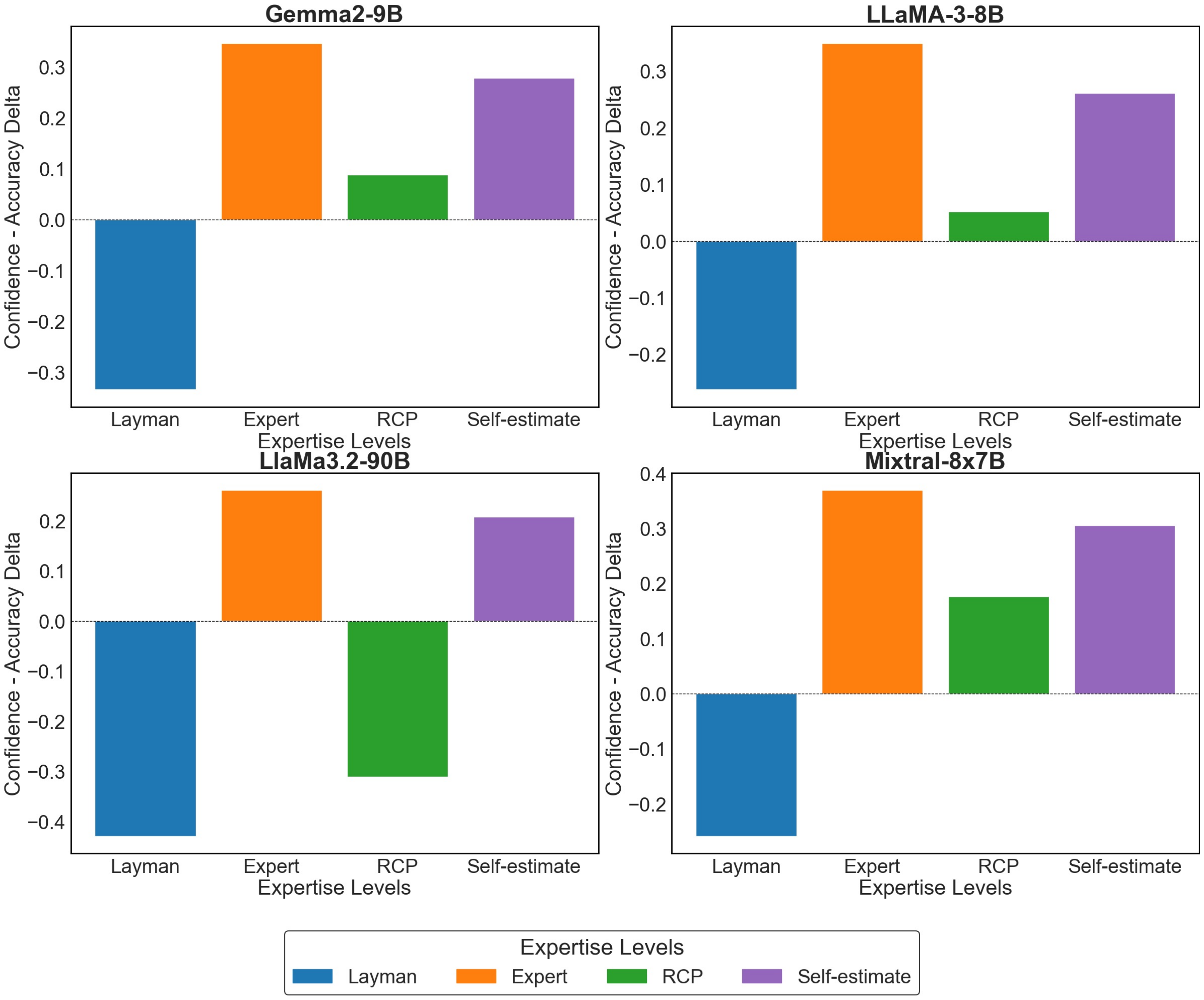}
    \caption{The difference between confidence and accuracy (Confidence - Accuracy) for three models across different personas and self-estimates, evaluated on physics, chemistry, and biology questions only. The top panel shows the main result, and the bottom panel shows the corresponding breakdown by role.}
    \label{fig:overplacement_combined}
\end{figure*}

\begin{figure*} 
    \centering
    \includegraphics[width=1\textwidth]{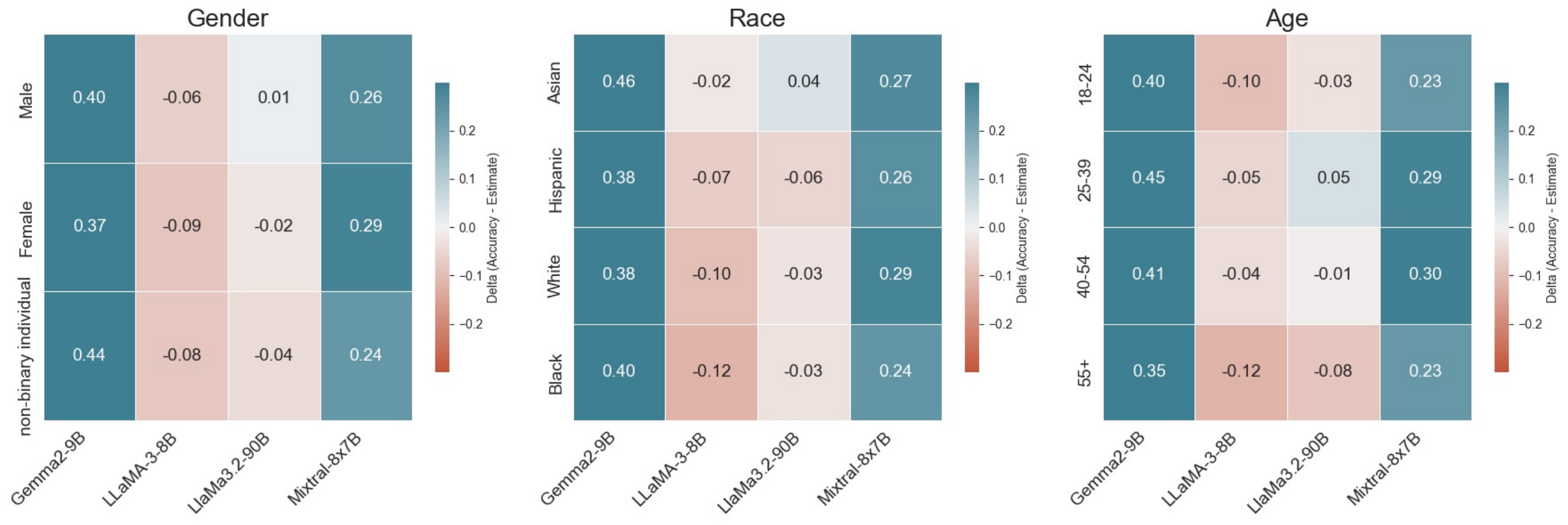} 
    \caption{$\Delta_\text{Demographic}$: The difference between confidence and accuracy ($\text{Confidence} - \text{Accuracy}$) for four models across gender, race, and age groups. Gemma2-9B and Mixtral-8x7B exhibit higher $\Delta_\text{Demographic}$ scores across all demographic categories, indicating greater bias. In contrast, LLaMA-3-8B and LLaMA-3-90B show substantially lower absolute values, suggesting better calibration and reduced demographic bias.}
    \label{fig:bias_results_1}
\end{figure*}

\subsection{Demographic Bias Results}
\label{fig:bias_results_1}

Figure~\ref{fig:bias_results_1} supplements our demographic bias experiments with results from four additional models. These results provide a more comprehensive view of how demographic attributes—such as gender, race, and age—impact model calibration and bias across different LLM architectures.

\begin{table*}[t!]
    \centering
    \scriptsize
    \setlength{\tabcolsep}{1.5pt}
    \begin{tabular}{@{}l*{27}{|c}@{}}
        \toprule
        \multirow{3}{*}{\textbf{Method}} & \multicolumn{9}{c|}{\textbf{High School}} & \multicolumn{9}{c|}{\textbf{College}} & \multicolumn{9}{c}{\textbf{Expert}} \\
        \cmidrule(l){2-28}
        & \multicolumn{3}{c|}{Physics} & \multicolumn{3}{c|}{Chemistry} & \multicolumn{3}{c|}{Biology} & \multicolumn{3}{c|}{Physics} & \multicolumn{3}{c|}{Chemistry} & \multicolumn{3}{c|}{Biology} & \multicolumn{3}{c|}{Physics} & \multicolumn{3}{c|}{Chemistry} & \multicolumn{3}{c}{Biology} \\
        \cmidrule(l){2-28}
        & Acc & AvC & ECE & Acc & AvC & ECE & Acc & AvC & ECE & Acc & AvC & ECE & Acc & AvC & ECE & Acc & AvC & ECE & Acc & AvC & ECE & Acc & AvC & ECE & Acc & AvC & ECE \\
        \midrule
        \multicolumn{28}{c}{\textbf{Gemma2-9B}} \\
        \midrule        
        Vanilla & 56.0 & 98.3 & 43.1 & 64.0 & 98.8 & 35.1 & \textbf{91.3} & 99.2 & \underline{7.9} & \underline{48.0} & 99.6 & 51.6 & 49.0 & 99.0 & 50.0 & \textbf{87.1} & 99.2 & \underline{12.1} & \underline{34.4} & 99.0 & 64.5 & \textbf{30.0} & 98.0 & 68.3 & \textbf{50.0} & 96.6 & 46.6 \\
        Top-K & 49.3 & 91.4 & 42.8 & 56.5 & 91.3 & 35.9 & 84.5 & 93.3 & 10.4 & 47.0 & 93.3 & 48.7 & 49.0 & 94.0 & 45.0 & 85.0 & 93.5 & \textbf{9.9} & 30.6 & 91.9 & 62.1 & 28.3 & 89.3 & \underline{62.7} & \underline{42.9} & 88.8 & \underline{46.0} \\
        Quiz-like & \textbf{58.7} & 96.0 & \underline{37.3} & \textbf{67.5} & 95.0 & \underline{27.5} & \underline{87.7} & 99.4 & 12.3 & \underline{48.0} & 93.0 & \underline{45.0} & \textbf{55.0} & 96.0 & \underline{41.0} & 86.4 & 99.3 & 12.9 & \textbf{36.7} & 92.8 & \underline{56.1} & 29.4 & 96.7 & 67.2 & 41.4 & 97.1 & 55.7 \\
        Ours & \underline{57.3} & 86.0 & \textbf{28.7} & \underline{66.0} & 90.0 & \textbf{24.0} & \underline{87.7} & 89.7 & \textbf{7.7} & \textbf{49.0} & 87.0 & \textbf{38.0} & \textbf{55.0} & 86.0 & \textbf{31.0} & \textbf{87.1} & 88.6 & 14.3 & \underline{34.4} & 71.7 & \textbf{37.2} & \textbf{30.0} & 79.4 & \textbf{49.4} & \underline{42.9} & 81.4 & \textbf{38.6} \\
        \midrule
        \multicolumn{28}{c}{\textbf{LLaMA-3-8B}} \\
        \midrule
        Vanilla & \textbf{43.3} & 83.3 & \underline{40.0} & \textbf{52.0} & 83.2 & \underline{31.7} & \textbf{79.7} & 84.6 & \textbf{5.5} & \underline{44.0} & 83.2 & 39.2 & \textbf{43.0} & 83.3 & 40.3 & \textbf{76.4} & 84.7 & 9.0 & 32.2 & 82.1 & 49.9 & 28.3 & 81.8 & 53.5 & \textbf{35.7} & 80.9 & 45.1 \\
        Top-K & 33.3 & 77.2 & 44.0 & 44.5 & 75.9 & \textbf{31.4} & 70.3 & 79.3 & 11.4 & \textbf{46.0} & 78.8 & \textbf{32.9} & \textbf{43.0} & 74.9 & \underline{35.3} & \underline{73.6} & 78.6 & \underline{8.2} & 28.3 & 71.2 & \underline{43.5} & 24.4 & 66.6 & \underline{42.2} & 30.0 & 70.5 & \underline{40.5} \\
        Quiz-like & 37.3 & 80.0 & 42.7 & \underline{47.5} & 80.5 & 33.0 & 73.5 & 85.5 & 11.9 & 41.0 & 83.0 & 42.0 & 42.0 & 83.0 & 41.0 & 72.9 & 85.0 & 12.1 & \textbf{34.4} & 81.1 & 46.7 & \underline{29.4} & 84.4 & 55.0 & 28.6 & 84.3 & 55.7 \\
        Ours & \underline{39.3} & 78.7 & \textbf{39.3} & 47.0 & 78.0 & 39.3 & \underline{73.9} & 80.0 & \underline{6.1} & 42.0 & 78.0 & \underline{36.0} & 42.0 & 72.0 & \textbf{30.0} & \underline{73.6} & 80.0 & \textbf{6.4} & \underline{33.9} & 53.3 & \textbf{19.4} & \textbf{31.1} & 61.1 & \textbf{30.0} & \underline{31.4} & 68.6 & \textbf{37.1} \\
        \midrule
        \multicolumn{28}{c}{\textbf{LLaMA-3.2-90B}} \\
        \midrule
        Vanilla & \textbf{65.3} & 92.4 & 27.8 & \underline{74.5} & 88.2 & 14.7 & \textbf{92.6} & 93.3 & \underline{3.3} & \underline{61.0} & 90.9 & \underline{29.9} & \textbf{58.0} & 85.8 & 28.8 & \textbf{95.7} & 91.3 & \underline{6.0} & 40.0 & 83.2 & 43.2 & \underline{33.9} & 81.6 & 47.8 & \underline{52.9} & 83.1 & \underline{30.3} \\
        Top-K & 60.7 & 88.4 & 28.7 & 73.5 & 79.7 & \underline{9.6} & 90.6 & 84.2 & 12.1 & \textbf{63.0} & 81.1 & \textbf{21.1} & 53.0 & 83.4 & 31.6 & 93.6 & 82.8 & 11.5 & 39.4 & 73.7 & \underline{35.4} & \textbf{35.6} & 68.6 & \textbf{34.5} & \textbf{55.7} & 69.6 & \textbf{18.1} \\
        Quiz-like & 61.3 & 86.7 & \underline{25.3} & 73.0 & 87.0 & 14.0 & \underline{91.9} & 97.4 & 6.1 & 48.0 & 83.0 & 35.0 & \underline{57.0} & 83.0 & \underline{26.0} & 93.6 & 95.7 & \textbf{2.1} & \textbf{42.8} & 81.1 & 38.3 & 30.6 & 83.3 & 52.8 & 47.1 & 93.3 & 37.1 \\
        Ours & \underline{62.7} & 80.0 & \textbf{17.3} & \textbf{75.0} & 80.0 & \textbf{5.0} & 80.3 & 80.6 & \textbf{1.0} & 49.0 & 80.0 & 31.0 & \underline{57.0} & 80.0 & \textbf{23.0} & \underline{95.0} & 80.0 & 15.0 & \underline{42.2} & 73.9 & \textbf{31.7} & 33.3 & 80.0 & \underline{46.7} & 44.3 & 80.0 & 35.7 \\
        \midrule
        \multicolumn{28}{c}{\textbf{Mixtral-8x7B}} \\
        \midrule
        Vanilla & \textbf{44.0} & 93.1 & 49.5 & \underline{54.5} & 91.3 & \underline{36.8} & \textbf{82.6} & 90.8 & \underline{8.3} & \textbf{51.0} & 91.6 & \underline{40.6} & \textbf{53.0} & 88.9 & \underline{35.9} & \textbf{84.3} & 89.5 & \textbf{6.0} & 32.8 & 87.5 & 54.7 & \underline{23.3} & 86.4 & 63.6 & \underline{44.3} & 84.5 & 40.8 \\
        Top-K & \underline{39.3} & 85.1 & \textbf{47.7} & 43.0 & 84.3 & 42.3 & 65.5 & 85.0 & 20.2 & 43.0 & 85.0 & 43.8 & 40.0 & 79.8 & 39.8 & 67.1 & 83.0 & 21.0 & 23.3 & 74.0 & \underline{51.8} & 22.2 & 67.5 & \textbf{45.8} & 31.4 & 69.8 & \underline{40.4} \\
        Quiz-like & \underline{39.3} & 90.0 & 50.7 & 34.5 & 86.5 & 52.0 & 25.2 & 82.9 & 57.7 & 25.0 & 85.0 & 60.0 & \underline{52.0} & 90.0 & 38.0 & 28.6 & 82.1 & 53.6 & \underline{35.0} & 89.4 & 54.4 & \textbf{25.0} & 89.4 & 64.4 & 27.1 & 81.4 & 54.3 \\
        Ours & 38.7 & 87.3 & \underline{48.7} & \textbf{59.0} & 90.0 & \textbf{31.0} & \underline{82.3} & 90.3 & \textbf{8.1} & \underline{47.0} & 85.0 & \textbf{38.0} & 46.0 & 81.0 & \textbf{35.0} & \underline{78.6} & 87.9 & \underline{10.7} & \textbf{36.1} & 70.6 & \textbf{34.4} & 20.0 & 68.3 & \underline{48.3} & \textbf{51.4} & 74.3 & \textbf{22.9} \\
        \bottomrule
        
    \end{tabular}
    \caption{Confidence elicitation and performance comparison for Gemma2-9B, LLaMA-3-8B, LLaMA-3.2-90B, Mixtral-8x7B on Physics, Chemistry, and Biology across three difficulty levels. Acc: Accuracy, AvC: Avg Confidence, ECE: Expected Calibration Error. All values are percentages. }
    \label{tab:method_compare_all_appendix}
\end{table*}

\subsection{LLM Prompt Templates}

\label{sec:prompt_templates}
This section details the exact prompt templates used in our experiments across various settings, including Answer-Free Confidence Estimation (AFCE), overplacement studies, and demographic bias analyses. Table~\ref{tab:dataset_examples} presents example questions at three different difficulty levels. Table~\ref{tab:hardness_prompts} provides a sample prompt for self-estimation, which is used in both the overconfidence and overplacement experiments. Additionally, Table~\ref{tab:overplacement_prompts} shows an example prompt used to test LLMs' confidence and accuracy when role-playing different personas. Finally, Table~\ref{tab:demographic_prompts} illustrates the prompt format used in demographic experiments involving gender, age, and race.

\begin{table}[h!]
\small
\begin{tabular}{p{0.9\linewidth}}
\toprule
// \emph{High school} \\
\textbf{Question}: A rigid, solid container of constant volume holds an ideal gas of volume v1 and temperature T1 and pressure P1. The temperature is increased in an isochoric process. Which of the following is NOT true? \\
A. The average speed of the molecules increases. \\
B. The pressure increases. \\
C. The kinetic energy of the system increases \\
D. The volume increases. \\
\textbf{Answer}: D \\
\midrule
// \emph{College} \\
\textbf{Question}: A uniform solid disk starts from rest and rolls down an inclined plane without slipping. After some time, what fraction of the disk’s total kinetic energy is rotational kinetic energy? \\
A. 1/4 \\
B. 1/3 \\
C. 1/2 \\
D. 2/3 \\
\textbf{Answer}: B \\
\midrule
// \emph{Expert} \\
In order to calculate the necessary beam to produce a particle X, we ran the following simulation in a High Energy software $e^{+}e^{-}\rightarrow X$, where the electron $e^{-}$ is at rest. We found that the beam energy has to be at least equal to $9.6\times10^{6}$ GeV. What is the mass of the particle X used for this simulation? \\
A. 1091 GeV \\
B. 99 GeV \\
C. 3132 GeV \\
D. 542 GeV \\
\textbf{Answer}: B \\
\bottomrule
\end{tabular}
\caption{Examples from dataset in three task hardness level.}
\label{tab:dataset_examples}
\end{table}

\begin{table}[h!]
\small
\begin{tabular}{p{0.9\linewidth}}
\toprule
// \emph{Prompt for accuracy calculation} \\
prompt = ```Please answer the following questions by selecting only the option letter. Do not provide any explanation. \\
Questions: \\
Q1: <> \\
Q2: <> \\
Q3: <> \\
Q4: <> \\
Q5: <> \\
Q6: <> \\
Q7: <> \\
Q8: <> \\
Q9: <> \\
Q10: <> \\
Now, please answer these questions with the following format: \\
   1. <Only Option Letter > \\
   2. <Only Option Letter > \\
   3. <Only Option Letter > \\
   4. <Only Option Letter > \\
   5. <Only Option Letter > \\
   6. <Only Option Letter > \\
   7. <Only Option Letter > \\
   8. <Only Option Letter > \\
   9. <Only Option Letter > \\
   10. <Only Option Letter > \\''' \\
\midrule
// \emph{Prompt for confidence calculation} \\
prompt = f"""Read the question, and estimate how many questions you can answer correctly (choose a number from 0-10). Do not provide any explanation. \\
Questions: \\
Q1: <> \\
Q2: <> \\
Q3: <> \\
Q4: <> \\
Q5: <> \\
Q6: <> \\
Q7: <> \\
Q8: <> \\
Q9: <> \\
Q10: <> \\
Estimate how many questions you can answer correctly: <choose a number from 0-10>.""" \\
\bottomrule
\end{tabular}
\vspace{2pt}
\caption{Prompts used for self-estimation experiments in our approach, independent of the AFCE method.}
\label{tab:hardness_prompts}
\end{table}

\begin{table}[h!]
\small
\begin{tabularx}{\linewidth}{p{0.9\linewidth}}
\toprule
// \emph{Prompt for accuracy calculation} \\
prompt = ```You are a \{role\}. Please answer the following questions by selecting only the option letter. Do not provide any explanation. \\
Questions: \\
Q1: <>\\
Q2: <> \\
Q3: <>\\
Q4: <> \\
Q5: <>\\
Q6: <> \\
Q7: <>\\
Q8: <> \\
Q9: <>\\
Q10: <> \\
Now, please answer these questions with the following format: \\
   1. <Only Option Letter > \\
   2. <Only Option Letter >  \\
   3. <Only Option Letter >   \\
   4. <Only Option Letter >\\
   5. <Only Option Letter >\\
   6. <Only Option Letter >\\
   7. <Only Option Letter >\\
   8. <Only Option Letter >\\
   9. <Only Option Letter >\\
   10. <Only Option Letter >\\''' \\
\midrule
// \emph{Prompt for confidence calculation} \\
prompt = ```A is an \{role\} in \{subject\}. Read the question, and after considering A’s ability, estimate how many questions A can answer correctly (choose a number from 0-10). Do not provide any explanation. \\
Questions: \\
Q1: <>\\
Q2: <> \\
Q3: <>\\
Q4: <> \\
Q5: <>\\
Q6: <> \\
Q7: <>\\
Q8: <> \\
Q9: <>\\
Q10: <> \\
Estimate how many questions you think A can answer correctly: <choose a number from 0-10>.''' \\
\bottomrule
\end{tabularx}
\vspace{2pt}
\caption{Prompts used for overplacement experiments in our approach, independent of the AFCE method.}
\label{tab:overplacement_prompts}
\end{table}

\begin{table}[h!]
\small
\begin{tabularx}{\linewidth}{p{0.9\linewidth}}
\toprule
// \emph{Prompt for accuracy calculation} \\
prompt = ```You are a \{race\} \{gender\} in the \{age\} age group. Read the question and provide your answer (only the option letter). Do not provide any explanation. \\
Questions: \\
Q1: <>\\
Q2: <> \\
Q3: <>\\
Q4: <> \\
Q5: <>\\
Q6: <> \\
Q7: <>\\
Q8: <> \\
Q9: <>\\
Q10: <> \\
Now, please answer these questions with the following format: \\
   1. <Only Option Letter > \\
   2. <Only Option Letter >  \\
   3. <Only Option Letter >   \\
   4. <Only Option Letter >\\
   5. <Only Option Letter >\\
   6. <Only Option Letter >\\
   7. <Only Option Letter >\\
   8. <Only Option Letter >\\
   9. <Only Option Letter >\\
   10. <Only Option Letter >\\''' \\
\midrule
// \emph{Prompt for confidence calculation} \\
prompt = ```You are a \{race\} \{gender\} in the \{age\} age group. Read the question, estimate how many questions you can answer correctly (choose a number from 0-10). Do not provide any explanation. \\
Questions: \\
Q1: <>\\
Q2: <> \\
Q3: <>\\
Q4: <> \\
Q5: <>\\
Q6: <> \\
Q7: <>\\
Q8: <> \\
Q9: <>\\
Q10: <> \\
Estimate how many questions you can answer correctly: <only choose one number from 0-10>.''' \\
\bottomrule
\end{tabularx}
\vspace{2pt}
\caption{Prompts for demographic experiments.}
\label{tab:demographic_prompts}
\end{table}

\section{Math Formula for ECE}
\label{appendix:ece_formula}
The Expected Calibration Error (ECE) is defined as:
\[
\text{ECE} = \sum_{m=1}^{M} \frac{|B_m|}{n} \left| \text{acc}(B_m) - \text{conf}(B_m) \right|
\]

Explanation of symbols:
\begin{itemize}
    \item $M$: The number of bins or groups into which predictions are divided.
    \item $B_m$: The set of predictions in the $m$-th bin.
    \item $|B_m|$: The number of predictions in the $m$-th bin.
    \item $n$: The total number of predictions across all bins.
    \item $\text{acc}(B_m)$: The accuracy of the predictions in the $m$-th bin.
    \item $\text{conf}(B_m)$: The average confidence of the predictions in the $m$-th bin.
\end{itemize}

\begin{table*}[ht]
    \centering
    \begin{tabular}{lcccccc}
        \toprule
        \textbf{Difficulty \& Subject} & \multicolumn{3}{c}{\textbf{Original Quiz-like}} & \multicolumn{3}{c}{\textbf{Original AFCE}} \\
        \cmidrule(r){2-4} \cmidrule(r){5-7}
        & \textbf{ECE} & \textbf{Confidence} & \textbf{Accuracy} & \textbf{ECE} & \textbf{Confidence} & \textbf{Accuracy} \\
        \midrule
        college\_biology & 5.0 & 97.1 & 95.0 & 3.6 & 97.1 & 95.0 \\
        college\_chemistry & 24.0 & 82.0 & 58.0 & 21.0 & 78.0 & 57.0 \\
        college\_physics & 18.0 & 88.0 & 70.0 & 17.0 & 88.0 & 71.0 \\
        gpqa\_biology & 21.4 & 80.0 & 58.6 & 10.0 & 72.9 & 65.7 \\
        gpqa\_chemistry & 30.6 & 69.4 & 38.9 & 16.1 & 51.1 & 37.8 \\
        gpqa\_physics & 32.2 & 70.6 & 40.6 & 21.1 & 57.8 & 41.1 \\
        high\_school\_biology & 4.8 & 99.7 & 94.8 & 6.1 & 99.4 & 94.5 \\
        high\_school\_chemistry & 12.0 & 92.5 & 80.5 & 9.5 & 89.0 & 81.5 \\
        high\_school\_physics & 12.7 & 86.7 & 74.0 & 14.7 & 88.0 & 73.3 \\
        \midrule
        \textbf{Difficulty \& Subject} & \multicolumn{3}{c}{\textbf{Random Order Quiz-like}} & \multicolumn{3}{c}{\textbf{Random Order AFCE}} \\
        \cmidrule(r){2-4} \cmidrule(r){5-7}
        & \textbf{ECE} & \textbf{Confidence} & \textbf{Accuracy} & \textbf{ECE} & \textbf{Confidence} & \textbf{Accuracy} \\
        \midrule
        college\_biology & 5.7 & 97.1 & 94.3 & 5.7 & 94.3 & 94.3 \\
        college\_chemistry & 24.0 & 82.0 & 58.0 & 14.0 & 74.0 & 60.0 \\
        college\_physics & 17.0 & 89.0 & 72.0 & 11.0 & 85.0 & 74.0 \\
        gpqa\_biology & 22.9 & 80.0 & 57.1 & 15.7 & 74.3 & 58.6 \\
        gpqa\_chemistry & 31.7 & 67.2 & 35.6 & 18.3 & 56.1 & 38.9 \\
        gpqa\_physics & 27.8 & 70.6 & 43.9 & 20.0 & 56.7 & 42.2 \\
        high\_school\_biology & 5.2 & 100.0 & 94.8 & 5.2 & 100.0 & 94.8 \\
        high\_school\_chemistry & 12.5 & 91.5 & 80.0 & 12.0 & 92.0 & 80.0 \\
        high\_school\_physics & 12.0 & 86.0 & 74.0 & 14.7 & 88.0 & 73.3 \\
        \bottomrule
        \textbf{Difficulty \& Subject} & \multicolumn{3}{c}{\textbf{5 Questions Quiz-like}} & \multicolumn{3}{c}{\textbf{5 Questions AFCE}} \\
\cmidrule(r){2-4} \cmidrule(r){5-7}
& \textbf{ECE} & \textbf{Confidence} & \textbf{Accuracy} & \textbf{ECE} & \textbf{Confidence} & \textbf{Accuracy} \\
\midrule
college\_biology & 6.4 & 100.0 & 93.6 & 6.4 & 98.6 & 95.0 \\
college\_chemistry & 38.0 & 96.0 & 60.0 & 22.0 & 78.0 & 56.0 \\
college\_physics & 28.0 & 99.0 & 73.0 & 27.0 & 96.0 & 69.0 \\
gpqa\_biology & 35.7 & 90.0 & 54.3 & 7.1 & 68.6 & 64.3 \\
gpqa\_chemistry & 26.7 & 65.0 & 38.3 & 23.9 & 60.0 & 36.1 \\
gpqa\_physics & 31.7 & 77.2 & 45.6 & 25.0 & 58.9 & 37.2 \\
high\_school\_biology & 4.5 & 100.0 & 95.5 & 4.8 & 100.0 & 95.2 \\
high\_school\_chemistry & 19.0 & 99.5 & 80.5 & 17.0 & 97.0 & 81.0 \\
high\_school\_physics & 21.3 & 96.7 & 75.3 & 22.0 & 96.7 & 74.7 \\
\bottomrule
    \end{tabular}
    \caption{AFCE remains effective regardless of question order or group size. Experiments with different question groupings (5 vs 10) and randomization showed no significant differences in outcomes, highlighting AFCE’s robustness across configurations.}
    \label{tab:random}
\end{table*}



\end{document}